\documentclass[journal]{IEEEtran}  
\ifCLASSINFOpdf
\else
\fi
\usepackage{subfiles}
\usepackage{standalone}
\usepackage{pdfpages}
\usepackage{wrapfig}

\def\idiapref{\textsuperscript{\textasteriskcentered}}
\def\epflref{\textsuperscript{\textdagger}}

\usepackage{graphics} 
\usepackage{epsfig} 
\usepackage{mathptmx} 
\usepackage{times} 
\usepackage{amsmath} 
\usepackage{amssymb}  
\usepackage{xspace}
\usepackage{paralist}
\usepackage{mathrsfs}

\usepackage{subfig}
\usepackage{makecell}
\usepackage{url}
\usepackage{colortbl}
\usepackage{multirow}

\newcommand{\mypartitle}[2][2.4]{\vspace*{-#1 ex}~\\{\noindent {\bf #2}}}

\newcommand{\NFeatureChannels}[0]{\ensuremath{N_w}\xspace}
\newcommand{\NStages}[0]{\ensuremath{S}\xspace}
\newcommand{\ConfidenceMapThreshold}[0]{\ensuremath{\eta}\xspace}
\newcommand{\PCKhDistanceThresh}[0]{\ensuremath{d}\xspace}
\newcommand{\PCKhCoef}[0]{\ensuremath{\kappa}\xspace}
\newcommand{\PCKhBBGtHeight}[0]{\ensuremath{h}\xspace}
\newcommand{\Features}[0]{\ensuremath{\mathbf F}\xspace}

\newcommand{\PartMaps}[0]{\ensuremath{\mathbf H}\xspace}
\newcommand{\LimbMaps}[0]{\ensuremath{\mathbf V}\xspace}
\newcommand{\StageIndex}[0]{\ensuremath{s}\xspace}

\newcommand{\SkeletonLoss}[0]{\ensuremath{L_{PM}}\xspace}
\newcommand{\DAMeasure}[0]{\ensuremath{L_d}\xspace}
\newcommand{\DALoss}[0]{\ensuremath{L_{DA}}\xspace}

\newcommand{\GFParams}[0]{\ensuremath{\theta_f}\xspace}
\newcommand{\GDParams}[0]{\ensuremath{\theta_d}\xspace}
\newcommand{\GYParams}[0]{\ensuremath{\theta_y}\xspace}

\newcommand{\Gf}[0]{\ensuremath{G_f}\xspace}
\newcommand{\Gy}[0]{\ensuremath{G_y}\xspace}
\newcommand{\Gd}[0]{\ensuremath{G_d}\xspace}

\newcommand{\DomainMixer}[0]{\ensuremath{\lambda}\xspace}
\newcommand{\DARegularizer}[0]{\ensuremath{R_d}\xspace}

\newcommand{\PartMapsLoss}[0]{\ensuremath{L_{\StageIndex}^{\PartMaps}}\xspace}
\newcommand{\LimbMapsLoss}[0]{\ensuremath{L_{\StageIndex}^{\LimbMaps}}\xspace}

\newcommand{\TeacherLoss}[0]{\ensuremath{L_{teacher}}\xspace}
\newcommand{\DistillLoss}[0]{\ensuremath{L_{distil}}\xspace}
\newcommand{\DistillStages}[0]{\ensuremath{L_{stages}}\xspace}
\newcommand{\DistillHints}[0]{\ensuremath{L_{hints}}\xspace}

\newcommand{\OurDataset}[0]{{DIH}\xspace}

\makeatletter
\DeclareRobustCommand\onedot{\futurelet\@let@token\@onedot}
\def\@onedot{\ifx\@let@token.\else.\null\fi\xspace}

\def\eg{{e.g}\onedot} 
\def\ie{{i.e}\onedot} 
 
\def\etc{{etc}\onedot}

\def\etal{{et al}\onedot}
\makeatother

\definecolor{Gray}{gray}{0.9} 

\begin{document}
%
\title{Efficient Convolutional Neural Networks for Depth-Based Multi-Person Pose Estimation}
%
%
%

\author{Angel Mart\'inez-Gonz\'alez\idiapref\epflref, Michael Villamizar\idiapref, Olivier Can\'evet\idiapref and Jean-Marc Odobez\idiapref\epflref
\thanks{Digital Object Identifier 10.1109/TCSVT.2019.2952779}
\thanks{URL: https://ieeexplore.ieee.org/document/8895819}
\thanks{$^{*}$ Idiap Research Institute, Switzerland. \{angel.martinez, michael.villamizar, olivier.canevet, odobez\}@idiap.ch }
\thanks{\textsuperscript{\textdagger} \'Ecole Polytechnique F\'ed\'erale de Lausanne (EPFL), Switzerland.}
}

%
%

\markboth{IEEE Transactions on Circuits and Systems For Video Technology }%
{Shell \MakeLowercase{\textit{et al.}}: Bare Demo of IEEEtran.cls for IEEE Journals}
%



\maketitle

\IEEEpubid{\begin{minipage}{\textwidth}\ \\[12pt] \centering
  \vspace{0.75cm}\\Copyright \copyright 2019 IEEE. Personal use of this material is permitted. 
  Permission from IEEE must be obtained for all other uses, in any current \\or future media, 
  including reprinting/republishing this material for advertising or promotional purposes,
 creating new collective works, \\ for resale or redistribution to servers or lists, or reuse of any 
copyrighted component of this work in other works
\end{minipage}}

\begin{abstract}
  Achieving robust multi-person 2D body landmark localization and pose estimation is essential for
  human behavior and interaction understanding as encountered for instance in HRI settings. 
  Accurate methods have been proposed recently, but
  they usually rely on rather deep Convolutional Neural Network (CNN) architecture,
  thus requiring large computational and training  resources.
  In this paper, we investigate different architectures and methodologies to address these
  issues   and achieve fast and accurate multi-person 2D pose estimation.
  To foster speed, we propose to work with depth images, whose structure
  contains sufficient information about body landmarks while 
  being simpler than textured color images and thus 
  potentially requiring less complex CNNs for processing. 
  In this context, we make the following contributions.
  i) we study several CNN  architecture designs combining pose machines relying on the cascade of detectors concept
  with lightweight and efficient CNN structures; 
  ii) to address the need for large training datasets with high variability,
  we rely on semi-synthetic data combining multi-person synthetic depth data 
  with real sensor backgrounds;
  iii) we explore domain adaptation techniques to address the performance
  gap introduced by testing on real depth images;
  iv) to increase the accuracy of our fast lightweight CNN models,
  we investigate knowledge distillation at several architecture levels 
  which effectively enhance performance.
  %
  Experiments and results on synthetic and real data  highlight
  the impact of our design choices, providing insights into
  methods addressing standard issues normally faced in practical applications,
  and resulting in architectures effectively matching our goal in both performance and speed.
\end{abstract}


\IEEEpeerreviewmaketitle


\section{Introduction}
\label{sec:intro}
\IEEEPARstart{B}{ody} landmark detection and human pose estimation are
fundamental tasks in computer vision.
They provide 
means for fine-level motion understanding
and human activity recognition, therefore finding applications
in many domains, \eg visual surveillance, gaming, social robotics,
autonomous driving, health care or Human Robot Interaction~(HRI).
However, real-time operation and computational budget constraints make
the deployment of reliable multi-person pose estimation systems challenging.

Recently, Convolutional Neural Networks (CNN) have become the
leading algorithms to address the human pose estimation task.
One trend is to design very deep models to achieve robustness to factors
like human pose complexity, self occlusion, scale, and noisy imaging.
A plethora of works have emerged proposing a very broad set of
CNN architectures configuration exploiting 
hierarchical features~\cite{yang2017pyramid,HourGlass,DeepPoseGoogle},
or human body relationships~\cite{Guler_2018DensePose_CVPR,CPMPaf}.
Generally, deeper architectures tend to perform better given their large
learning capacity.
Yet, their excellent performance comes with the drawback of requiring large computational 
resources, since state-of-the-art designs comprise millions of parameters.
In addition, their operation is hindered by the low budget GPU devices normally available 
in practical applications.
This paper focuses on fast and reliable 2D multi-person pose estimation.
We investigate methods addressing specific subgoals of our overall task, \ie 
training data availability, efficiency and performance.
First, we propose to rely on depth images.
Contrary to RGB images, which have a high diversity of color and texture content,
a depth image contains mainly shape information making them relatively 
simpler while still containing rich and sufficient information for human
pose estimation, as shown by Shotton \etal \cite{ShottonPAMI}.
They may thus necessitate less complex models to accomplish the task.
Although the depth information would allow to address the 3D pose estimation case, 
in this paper we focus on 2D pose in the multi-party case as we believe it is 
an important task that can be used in practical applications and can be a
valuable first step towards 3D pose once individuals have been detected.
Secondly, we investigate efficient CNN architectures  with good speed-accuracy trade-off 
inspired from recent advances in efficient convolution structures 
and designed to operate in real-time with low computational resources.
Third, to address the need for training data, we rely on a semi-synthetic approach
in which multi-person depth data are rendered using a randomized rendering pipeline,
and merged with real backgrounds corresponding to the employed depth sensor.
To address the resulting domain gap, we investigate domain adaptation techniques,
showing that fine-tuning is very efficient, while domain adversarial adaptation
is not really working on our data.
Finally, we explore knowledge distillation as a way to increase the generalization
performance of our lightweight models.
These ideas are motivated in the following paragraphs.

\vspace*{-0.15cm}
\subsection{Motivations}

%
CNN-based human pose estimation methods traditionally use a deep architecture
pretrained on a large scale image recognition dataset.
%
This design choice might unnecessary bring high computational burden.
In this paper, inspired by efficient network structures such as those encountered in 
ResNets~\cite{He_RESNET_CVPR_2016},
MobileNets~\cite{Mobilenets_Howard_2017} and
SqueezeNets~\cite{Squeezenet_Iandola_2016},
we introduce novel lightweight network architectures that match our real-time
and performance requirements.
Our designs adopt the multi-stage prediction scheme to sequentially refine
predictions as in a cascade of detectors approach.
Nevertheless, as training these  lightweight models from scratch using only the ground truth 
annotations might be harder due to their lower learning capacities, 
we boost their generalization abilities by employing
knowledge distillation~\cite{Hinton_NIPS_2015}.
Knowledge distillation aims to transfer the generalization capacities
of larger and more accurate models to smaller and more efficient models.
We employ different distillation techniques and illustrate how to couple
these with our architectures designs to improve performance while maintaining 
efficiency.

A critical element for any CNN-based approach is to have a large and varied 
dataset for the learning stage.
In the human pose estimation literature some common datasets contain thousands of images 
properly labelled by human annotators \cite{MSCOCO_2014}.
Yet, large quantities of labelled data are not always available for specific application 
settings or are  expensive to produce.
Therefore, researchers have also investigated the use of computer graphics to synthesize 
large datasets for applications requiring a different set of annotations, \eg
human 3D pose estimation \cite{Fabbri_ECCV_2018,Chen_3DVision_2016} or
3D character manipulation \cite{UnitePeople_Lassner_2017_CVPR}.
%
The key element with synthetic images is that  annotations
come at no cost since these are generated during rendering.

As there are no large datasets of depth images with proper body landmark annotations
for training, we propose to rely on semi-synthetic images.
Compared to color images, synthesizing depth images is easier due to
their independence to color, material, visual texture or lighting conditions.
Nonetheless, relying only on synthetic  images for learning will result in
a performance drop when testing with real depth images,
since synthetic and real images present large differences in 
their visual characteristics.
One typical example is the noise due to depth discontinuities in real sensors
which  differs significantly from the noise free synthetic data.
To reduce the  performance drop, we investigate several approaches.
The first one is to fuse synthetic depth images of persons (for which the annotation is known)
with background depth data from the real sensor, generating semi-synthetic data allowing the network
to already learn sensor noise characteristics.
A second approach is to use unsupervised adversarial domain adaptation. 
It is a framework  enabling learning a body landmark detector using only annotations
on synthetic images while leveraging the information contained in large quantities of unannotated real ones.
The main idea is to split the 2D CNN pose predictor in two parts, 
a feature extractor and an  actual body landmark localizer,
and to train them
so that the localization performance on the synthetic data are good,
while at the same time, it should confuse a domain classifier trained
at identifying the  domain (where the data are coming from, synthetic or real) of the extracted features,
see Fig.~ \ref{fig:res-arch}.

Note that learning body landmark detection models with synthetic depth images has been
already proposed in the literature~\cite{ShottonPAMI}.
However, in our case we focus on a more general image synthesis approach by
considering simulations at far and close ranges, involving multiple people with occlusions,
and fusing synthetic with real data.
In addition, both our synthetic and real datasets
are made publicly available upon~request\footnote{\url{https://www.idiap.ch/dataset/dih}}.

Finally,
given the unavailability of benchmark datasets for 2D multi-person pose estimation 
from depth images, we have collected a set of video sequences with a Kinect~2 sensor.
The sequences simulate multi-person HRI scenarios with a humanoid robotic platform,
with people under different degrees of occlusion and distance from the sensor.

\vspace*{-0.15cm}
\subsection{Contributions and paper outline}

This paper addresses efficient and reliable multi-person body landmark detection
and pose estimation.
We propose to rely on depth data and on lightweight CNN architectures
suitable to achieve a good accuracy-efficiency trade-off.
To alleviate the need for manual annotations, we propose to use synthetic
depth images for training.
We bridge the performance gap provoked by
learning from synthetic images and testing on real ones by employing domain
adaptation techniques.
Finally, we illustrate the use of knowledge distillation methods to boost
the performance of lightweight models.
In that context, our contributions can be summarized as follows:

\begin{itemize}
	\item we propose to use depth data for robust human body landmark localization,
	and semi-synthetic depth images for the learning stage;
	\item we investigate different lightweight CNNs architectures comprising a cascade of detectors and 
	inspired from ResNets, MobileNets and SqueezeNets efficient designs;
	\item we investigate the use of adversarial domain adaptation
	of neural networks for our body landmark detection task and report
	its limitations;
	\item we explore knowledge distillation techniques and show how to 
	couple them with our CNN designs to boost the generalization
	abilities of our novel lightweight models.
\end{itemize}

In~\cite{Martinez_IROS_2018} we had introduced a residual pose machine architecture,
while in~\cite{Martinez_ECCVW_2018} we had investigated the adversarial adaptation 
scheme.
This paper extends these previous works in several ways: we introduce and analyze
new faster CNN architectures, based on MobileNets and SqueezeNets;
we propose and show that knowledge distillation is an effective way to improve the performance of these
lightweight models.
In addition, we include more experimental validation and analysis including an additional
comparison with the state-of-the-art and experiments with another public dataset.

\vspace*{1mm}

The reminder of this paper is organized as follows.
Section~\ref{sec:state-of-the-art} presents an analysis of the state of the art on
human pose estimation, domain adaptation and knowledge compression.
Section~\ref{sec:data-domains} presents the synthetic and
real depth image databases used in our experiments.
Section~\ref{sec:netdetails} introduces our efficient CNN.
Adversarial domain adaptation and knowledge distillation
are introduced in Sections~\ref{sec:domain-adaptation} and \ref{sec:distillation}.
Experimental protocol and results are presented in Section~\ref{sec:experiments},
and Section~\ref{sec:conclusions} concludes the work.

\section{Related Work}
\label{sec:state-of-the-art}
%
This section presents an overview of the literature concerning our fast 
human pose estimation task with CNNs.
In addition, we briefly review recent advances in the domain adaptation literature
and on knowledge distillation.

\vspace{-0.15cm}
\subsection{Human Pose Estimation}
Human pose estimation has been a computer vision subject studied for decades.
Classical machine learning methods employed body part specific detectors 
over hand crafted features to model body landmark relationships in 
tree-like structures~\cite{Yang_PAMI}.
More recently, as with other computer vision tasks, CNNs have become the dominant approach.
They address the single-person body landmark localization task by
enlarging the CNN receptive field~\cite{DeepCut},
or combining features from inner architecture layers 
\cite{HourGlass,yang2017pyramid}.
The Multi-person case can then be addressed by using an additional person 
detector at the expense of a larger computational cost.

To move one step further and directly solve the multi-person pose estimation problem, spatial 
and temporal relationships between pairs of body parts have been either modeled
by explicit relationship predictors \cite{ArtTrack, PoseTrack}, or embedded in the CNN
architecture~\cite{CPMPaf,Fabbri_ECCV_2018}.
For example, the work of Cao \etal \cite{CPMPaf} has successfully applied the
concept of cascade of detectors to build a CNN architecture that
extracts features from the image and then relies
on a sequence of stacked layers to refine predictions and encode body
parts pairwise dependencies.
%
%
%
%
Recent works applying this concept with CNNs have proposed to model detectors 
as groups of convolutional layers~\cite{HourGlass} or recurrent 
modules~\cite{ErrorFeedback,Luo_LSTM_CVPR_2018}.
%
%
Yet, the more detectors, the higher the need for computational resources.

Depth data has also been used for human pose estimation related tasks.
%
%
%
%
Depth-based features have been successfully used to build body part classifiers
via random forests and trees~\cite{ShottonACM,Jung_RTW_CVPR_2015} or
support vector machines~\cite{DepthGestureSVM}.
%
%
The seminal work of Shotton \etal~\cite{ShottonACM} 
uses a random forest based on 
simple depth features to label pixels as one of the different body parts.
%
%
Despite  remarkable results, the method assumes 
background subtraction as a preprocessing step, and is limited to 
near-frontal pose and close-range observations.

CNN-based methods have also been proposed for articulated human 
pose estimation from depth images
\cite{DepthInvariant, DepthMultitask,Moon_2018_CVPR_V2V-PoseNet}.
The approaches in~\cite{DepthInvariant, DepthMultitask} address body landmark detection as a patch
classification task for single person pose estimation.
%
For example, \cite{DepthInvariant} solves the problem of multi-view
pose estimation setting by learning a view invariant feature space 
from local body part regions.
3D coordinate estimations of body landmarks are provided by a Recurrent
Neural Network (RNN) in an error feedback mechanism to correct estimates.
Yet, besides requiring iterations to produce a final estimation, the 3D coordinate
regression mechanism is highly non-linear.
The approach in~\cite{Moon_2018_CVPR_V2V-PoseNet} 
converts depth images to voxels and compute the likelihood of each body landmark per voxel.
This avoids the large non-linearity but introduces a heavy 3D-CNN for voxel
processing, on top to the heavy depth image preprocess to generate voxels.
%
%

%
In our work, in contrast with these previous methods, we predict 2D landmark locations
in the depth image without requiring a heavy image preprocessing.
In addition, we apply the cascade of detectors concept
to depth images, while at the same time exploring more efficient network architectures
based on ResNet modules.

\vspace{-0.15cm}
\subsection{Domain Adaptation}
%
%
The premise of a domain adaptation technique is to learn a data 
representation that is invariant across domains.
A good analysis of the state-of-the-art is given in~\cite{DASurvey}.
Current deep learning methods
employ specialized CNN architectures to learn an invariant representation
of the data
via adversarial learning~\cite{Ganin_DANN},
residual transfer~\cite{NIPS2016_residual_transfer_long},
and Generative Adversarial Networks (GAN)~\cite{Hoffman_CYCLEDA_ICML_2018}.
For example, \cite{Ganin_DANN} proposes
a discriminative approach to learn domain invariant features for object classification.
More precisely, the method jointly learns an object class and domain predictors 
relying on shared features between the two tasks.
In this context, domain adaptation is achieved by minimizing the object
recognition loss while maximizing the error on domain classification.
Other recent approaches like \cite{Hoffman_CYCLEDA_ICML_2018}
seek to adapt the data representations at pixel level, in addition to
the feature space,
%
%
enforcing as well consistency of specific semantic features
to detect objects in both domains.

Deep domain adaptation for depth images is less covered in the literature.
The work in \cite{DepthDA_Patricia_ICCV} 
compares domain adaptation techniques applied to object classification from depth images.
It shows that there is an intrinsic difficulty in performing
adaptation given that noise in depth images is significantly more
persistent and different between sensors as compared to that among RGB images.
%
%
%

To the difference of most domain adaptation works, which focus in object classification tasks
in settings with few domain differences (objects perspective, image background 
and lighting),
we analyse unsupervised domain adaptation for a regression task.
Additionally, we contrast its limitations with a simple finetuning approach.


\begin{figure*}[tb]
	\center
	\includegraphics[width=.98\linewidth]{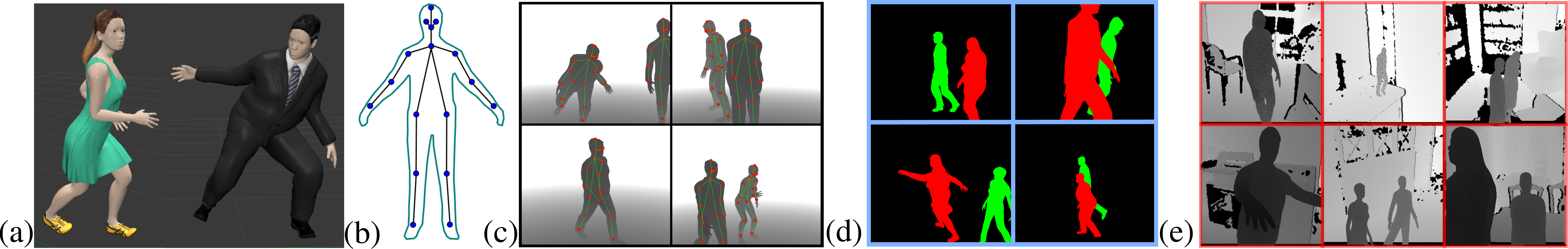}
	\vspace{-1mm}
	\caption{(a) Sample 3D characters with different poses and outfits;
	(b) skeleton model;
	(c) rendered synthetic depth image sample;
	(d) examples of generated colored depth mask for synthetic images with more than one person;
	(e) examples  of training images, combining synthetic generated 
	bodies with real background images.
	}
 \vspace{-2mm}
\label{fig:randomscenario}
\end{figure*}

\vspace{-0.15cm}
\subsection{Model compression and knowledge distillation}

Accurate CNN rely in increasingly deeper architectures.
Normally, these are over-parameterized and binded with large computational
cost, hindering their practical application.

Recent works overcome this over-parametrization by exploring channel 
pruning~\cite{Molchanov_Pruning_ICLR_2016}
or by designing lightweight CNN architectures~\cite{Mobilenets_Howard_2017,Squeezenet_Iandola_2016}.
On one hand,
model pruning remove filters
that produce statistically very low activations during the learning process.
On the other hand, lightweight architectures designs aim at directly
increasing speed by exploiting different convolution strategies.
For example, the MobileNet \cite{Mobilenets_Howard_2017}
factorizes standard convolutions into depthwise and pointwise convolutions,
reducing computation and model size.
The SqueezeNet model~\cite{Squeezenet_Iandola_2016} constrains the number
of input feature channels of layers with large kernel size.
These architecture designs have made a tremendous gain towards efficiency,
with some of them being able to run 50 times faster than their deeper 
counterpart with small loss in performance.

Another line of research that pursues the same goal is knowledge distillation
\cite{Hinton_NIPS_2015,Romero_Hints_ICRL_2015,Suraj_ICML_2108,Chen_NIPS_2017}.
Its objective is to train a small and light CNN, called the \emph{student}
model to mimic a more complex and accurate model named the \emph{teacher}.
The pioneering work in \cite{Hinton_NIPS_2015}
showed how the student model can acquire the teacher knowledge
using the teacher activations as soft targets.
%
%
%
%
To boost the student's generalization capabilities, recent works \cite{Romero_Hints_ICRL_2015}
have proposed to introduce "hints" as an attempt to additionally mimic the activations 
of a given hidden layer of the teacher.
Matching the teacher and student Jacobians of the learning objective
has also been used during distillation \cite{Suraj_ICML_2108}
and can be seen as a form of data augmentation with Gaussian noise.
%
%
In \cite{Chen_NIPS_2017} knowledge distillation is applied to learn an 
efficient model for multi-object detection and bounding box regression.
The approach exploits the teacher
output as an 
upper bound
and apply penalty only when the
desired output is below a certain margin.

A novelty of our work is to apply knowledge distillation for body landmark detection
and a regression task.
In doing so, we strongly couple distillation with the process of refining body landmark predictions
and show this improves the performance of our efficient architectures compared to simpler
knowledge distillation or learning with hints.

\section{Depth Image Datasets}
\label{sec:data-domains}
In this section we introduce the datasets of depth images we used
for training and testing.
%
%
We start by describing the methodology used to generate our synthetic image dataset.
Then, we describe the dataset of real depth images collected with a Kinect~2
sensor.

\vspace{-0.15cm}
\subsection{Synthetic depth images dataset}
The appropriate training of CNNs requires large amounts of data with
high quality ground truth.
Unfortunately, building datasets of humans with annotated body landmarks at
large scale can be very expensive.
%
%
%
We overcome this by relying on computer graphics and
considering the \OurDataset synthetic depth 
image database introduced in \cite{Martinez_ECCVW_2018}.
The dataset contains images displaying single and two people instances
with different body pose and view perspectives.

The synthetic dataset was generated by a randomized rendering pipeline.
We used real motion capture data (mocap) to perform motion retargeting to 
a 3D character and applied variations in viewpoints to generate synthetic
images with body landmark ground truth.
The challenge is how to automatically introduce variability in human shapes, 
body pose and view point configurations. This is detailed below.

\mypartitle{Dataset and annotations.}
The dataset contains 264,432 images of people performing different 
types of motion under different viewpoints with 71,711 images displaying 
two people. Some examples are shown in Figure~\ref{fig:randomscenario}(c).
The body skeleton comprises 17 body landmarks as shown in 
Figure~\ref{fig:randomscenario}(b).
The landmarks, \ie \emph{head, neck, shoulders, elbows, wrists, 
hips, knees, ankles, eyes}, are extracted in 3D camera and 2D image coordinates.
The dataset also provides keypoint visibility and color labeled silhouette  masks
%
which can be used to determine keypoint visibility and to perform
data transformations during training like adding pixel noise or fusing with real backgrounds
(see below).
See Figure~\ref{fig:randomscenario}(d) for some examples.

\mypartitle{Variability in body shapes}.
We rely on a dataset of 24 3D characters that show variation in 
gender, heights and weights, and were dressed with 
different clothing outfits to increase shape variations
(skirts, coats, pullovers, \etc).
See Figure~\ref{fig:randomscenario}(a).

\mypartitle{Synthesis with two people instances}.
Multi-person pose estimation scenarios are covered by adding 
two 3D characters to the rendering scene. During synthesis, two 
models are randomly selected from the character database and
placed 
%
randomly in the virtual scene, but keeping a minimum distance
between them to avoid checking for collision.

\mypartitle{Variability in body poses}.
Motion simulation is used to add variability in body pose  
configurations.
We performed motion retargeting from motion capture data sequences 
taken from the CMU Mocap dataset~\cite{CMUMocap}.
Amongst the available actions, we selected  the following ones as most representative of our target scenario:
walking, jogging, jumping, stretching, turning and various sport gesturing.

\mypartitle{Variability in view point}.
A camera is randomly positioned at a maximum distance of 8 meters 
from the models, and randomly oriented towards the models torso.

\mypartitle{Real background fusion}.
%
%
A solution would be to generate random background content by randomly placing
3D objects in the rendering pipeline.
This is however non-trivial as a large variety of objects is needed to cover
all expected backgrounds.
For this reason, in this paper we propose instead to use background images from the
target depth sensors.
This has 2 main advantages.
First, a large variety of background scenes with different contents can easily 
be collected.
Second, the resulting data will already contain sensor specific information
that will aid in the generalization capabilities of the learned models.
In practice, background generation is performed on the fly during training
by fusing the real background image content with the synthetic depth images with people.
Some example images are shown in Figure~\ref{fig:randomscenario}(e).
Section~\ref{sec:synthetic-training} describes this process in detail.

\mypartitle{Body landmark visibility labels}.
We provide visibility labels for each of the body landmarks in the image.
The visibility labels account for partial and self occlusions.
%
A body landmark is set to visible by thresholding the distance between the landmark
and the body surface point that projects onto the image at the same position than the landmark.
%
%
%
%

\vspace{-0.15cm}
\subsection{Real depth image dataset}
\label{sec:real-domain-data}

\begin{figure}
	\center
	\includegraphics[width=\linewidth]{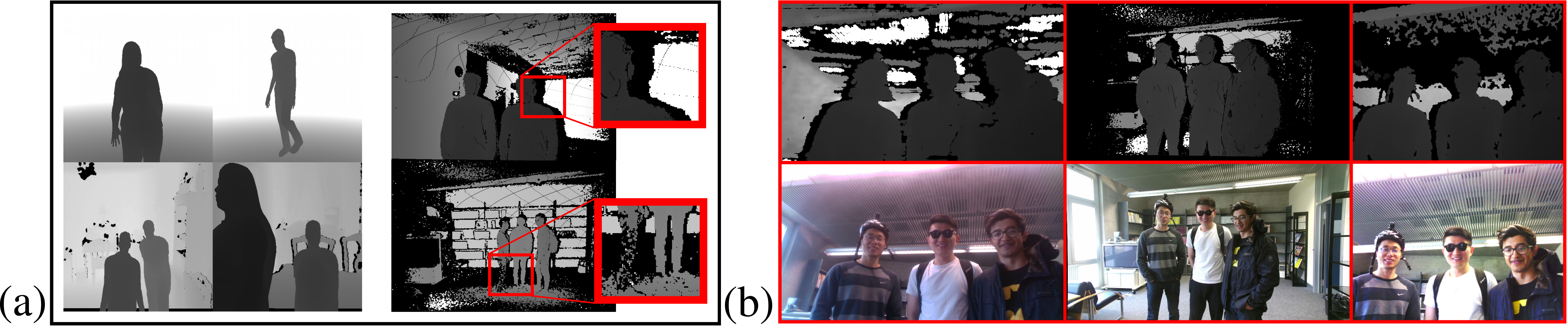}
	\vspace{-4mm}
	\caption{Depth imaging characteristics.
	  (a)
	  some visual characteristics of real depth images (right)
	  like  shadows around the silhouette or sensing failures 
	  due to surface material and depth variation (red square)
	  are difficult to synthesize and therefore not present in  the synthetic images (left);
	  (b) HRI scene recorded with different RGB-D cameras.
	  Left to right: Intel D435, Kinect~2 and Asus Xtion.
	  Different depth sensors have different quality characteristics.
	}
          \vspace{-2mm}
\label{fig:sensing-types}
\end{figure}

Depth imaging is usually the result of a triangulation process in which a 
series of laser beams are cast into the scene, captured by an 
infrared camera, and correlated with a reference  pattern to 
produce disparity images and finally the distance to the sensor.
As a result, the image quality  greatly depends on the 
sensor specifications like  measurement variance, missing data, 
surface discontinuities, etc.
%
%
%
Samples from three different sensors are shown in Figure~\ref{fig:sensing-types}(b).
%
%
In particular, sensor limitations such as shadows around the silhouette or 
sensing failures  due to surface material and depth variations are not present in the
synthetic images since they  are difficult to realistically simulate.
As a result and as illustrated in Figure~\ref{fig:sensing-types}(a),
there exists a large difference in the visual features exhibited
by synthetic and real depth images.

In this paper we relied on data from the Kinect~2 sensor.
Compared to other sensors like Intel D435 or Asus Xtion, 
it has a more accurate depth estimation and a large
field of view which is better for HRI analysis.
We consider the real data fold of the \emph{\OurDataset} dataset.
It contains 16 indoor sequences of up to three minutes composed of pairs of registered color and depth images.
They display up to three people captured  at different distance from the
sensor and with different levels of occlusion and scene backgrounds.
A total of 9 different participants were involved
in natural HRI interaction situations (walking off and towards the robot,
stretching hands and between person interactions).
They wore different clothing accessories to add variability 
in body shape.
%

\section{Efficient human pose estimation}
\label{sec:netdetails}

This section describes the efficient CNN models we investigated for the task of body
landmark localization and pose estimation.
We start by introducing the pose machine architecture 
which forms the base of all our models.
Then, we present the different instantiations that we investigated to
improve efficiency while maintaining accuracy.

\vspace{-0.1cm}
\subsection{Pose machines CNN architecture class}
\begin{figure}
\center
\includegraphics[width=1\linewidth]{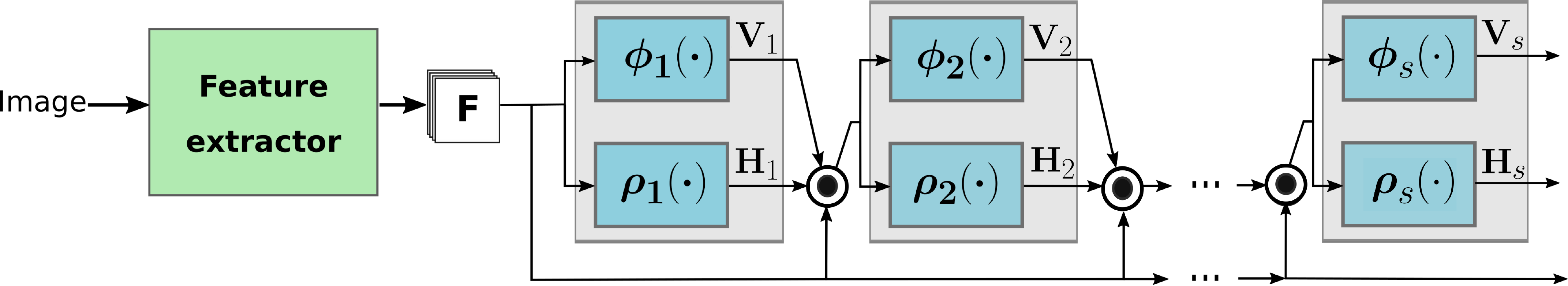}
\vspace{-0.6cm}
\caption{Pose machine architecture class. It comprises a feature extractor
		 module and a prediction cascade.
		 Each stage in the prediction cascade is composed by two branches that
		 predict confidence maps of body landmarks and body limbs in the image.
		 %
		 %
		 They take as input the extracted features $\mathbf \Features$ and 
		 the confidence maps from the previous stage to refine predictions.
}
\label{fig:cpm-sketch}
\end{figure}

The pose machine architecture comprises two main
components: a feature extractor module and a cascade of predictors that output
confidence maps for each of the body landmarks and body limbs.
Figure~\ref{fig:cpm-sketch} sketches the architecture class concept and its main
components.

More precisely, the CNN takes an image as input and the feature
extractor module computes an abstract representation of
it composed of $\NFeatureChannels$ channels, denoted as $\Features$.
These features $\Features$ are passed to the cascade of predictors, composed of a
series of prediction stages sequentially stacked.
Each prediction stage aims at localizing body landmarks (neck, elbows, ankles) and limbs,
which are segments between two landmarks according to the skeleton shown in
Figure~\ref{fig:randomscenario} (forearms or thighs).

Each stage $\StageIndex$ consists of two branches made of fully convolutional
layers predicting confidence maps of body landmarks, denoted $\rho_{\StageIndex}(\cdot)$, and
of body limbs, denoted $\phi_{\StageIndex}(\cdot)$.
For $\StageIndex\geq 2$ these branches take as input both the features $\Features$ and the landmark
and limbs predictions maps from stage $\StageIndex-1$.
In effect, this allows the refinement of the predictions of each element (landmark and limbs) by
incorporating context from the other body parts and hence accounting for valid body pose configurations.
This effectively reduces the number of pairs of detected body landmarks and potentially connected,
easing the single and multi-person pose inference stage.

 \begin{figure}
 \centering
 \includegraphics[height=1.3\linewidth]{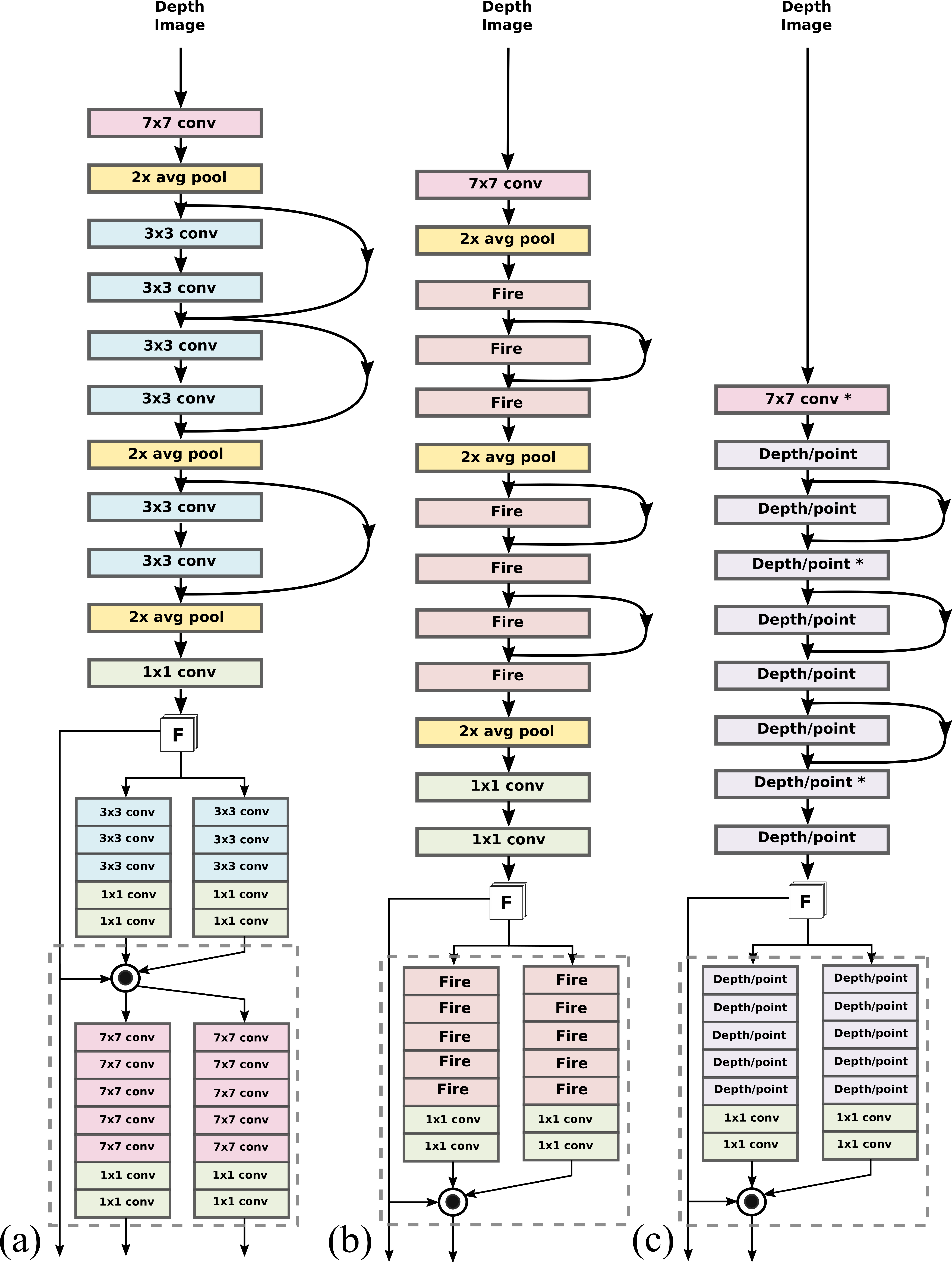}
 \vspace{-0.2cm}
 \caption{Architecture instances of the pose machines class.
 		 (a) Residual pose machines focuses on speeding up the feature
 		 extractor module using ResNet modules;
 		 (b) SqueezeNet pose machines builds on the \emph{Fire} module concept
 		 to design a lighter architectures;
 		 (c) MobileNet pose machines relies on depthwise and pointwise
 		 convolution layers to speed up computation.
 		 Convolution layers marked with * have a stride of 2 and serve as 
 		 pooling mechanism.}
 \label{fig:net-archs}
 \end{figure}

\vspace{-0.1cm}
\subsection{Efficient pose machines}

For an efficient forward pass, instances of the above architecture class will
incorporate lightweight designs in the feature extractor, $\rho_{\StageIndex}(\cdot)$ and $\phi_{\StageIndex}(\cdot)$.
In Figure~\ref{fig:net-archs} we illustrate the design of our efficient pose machine instances.
%
Modules enclosed by doted squares are the components which are replicated to achieve
the cascade of predictors.
We describe these architectures in the following.

\subsubsection{Residual pose machines}
In the pose machine architecture instance presented in \cite{CPMPaf}, the
first computational bottleneck is the
large VGG-19 architecture used as feature extractor module.
Therefore, we propose to investigate how to exploit a lighter module built
upon residual modules (or blocks)~\cite{He_RESNET_CVPR_2016}.
We originally introduced this modification in~\cite{Martinez_IROS_2018}.
Our motivation is that residual blocks are known to outperform VGG
networks, and to be faster by having a lower computational
cost~\cite{canziani2016analysis}.
Figure~\ref{fig:net-archs}(a) depicts the architecture we dubbed as \emph{residual
pose machines}.
%
%

\mypartitle{Feature extraction network.}
It consists of an initial convolutional layer followed
by three residual modules with small
kernel sizes (${3 \times 3}$).
The network has three average pooling layers.
Each residual module consists of two
convolutional layers and a shortcut connection.
%
Batch normalization and ReLU are included after all convolutional layers and
shortcut connections as exemplified in Figure~\ref{fig:zoom-in}(a).

\mypartitle{Pose regression cascade.}
We maintain a large effective receptive field in the design of the branches
$\phi_{\StageIndex}(\cdot)$ and $\rho_{\StageIndex}(\cdot)$.
In the first prediction stage the network has three convolutional layers
with filters  of ${3 \times 3}$ and two layers with filters of ${1 \times 1}$, whereas
in the remaining stages there are five and two convolutional layers with
filters of ${7 \times 7}$ and ${1 \times 1}$ respectively.

\subsubsection{Squeezenet pose machines}
In~\cite{Squeezenet_Iandola_2016} the SqueezeNet architecture is build
upon a series of modules called \emph{Fire} modules.
Each module is composed by a \emph{squeeze} layer and an \emph{expand} layer.
The squeeze layer contains filters of $1\times 1$ and outputs $N_{s_1}$ channels,
while the expand layer is a mix of filters of ${1 \times 1}$ and ${3 \times 3}$ that outputs $N_{e_1}$ and
$N_{e_3}$ channels respectively.
This configuration aims to reduce the model size by using
$1\times 1$ filters, and to speed up computation by
limiting the number of input feature channels of the $3\times 3$ layers.
%
%
%
Figure~\ref{fig:zoom-in}(b) illustrates the composition of a Fire module.
%

Our architecture design using Fire modules is shown in Figure~\ref{fig:net-archs}(b).
The output of the Fire module is the concatenation of the channels from
the $1\times 1$ and $3\times 3$ expand layers.
We follow the original design and set $N_{s_1}$ to be less than
the sum $N_{e_1} + N_{e_3}$ and maintain these quantities low.
%
Batch normalization and ReLU are added after each convolution.

\mypartitle{Feature extraction network.}
The feature extractor module comprises 7 Fire modules, three average
pooling layers, three residual connections and two $1\times 1$ convolution layers.
ReLU activations are included after the
shortcut connection following the design in Figure~\ref{fig:zoom-in}(a).

\mypartitle{Pose regression cascade.}
The pose regression cascade employs five Fire modules and two $1\times 1$ convolution layers 
in the design of both branches $\phi_{\StageIndex}(\cdot)$ and $\rho_{\StageIndex}(\cdot)$.

\subsubsection{Mobilenet pose machines}
\begin{figure}[t]
 \centering
 \includegraphics[width=1\linewidth]{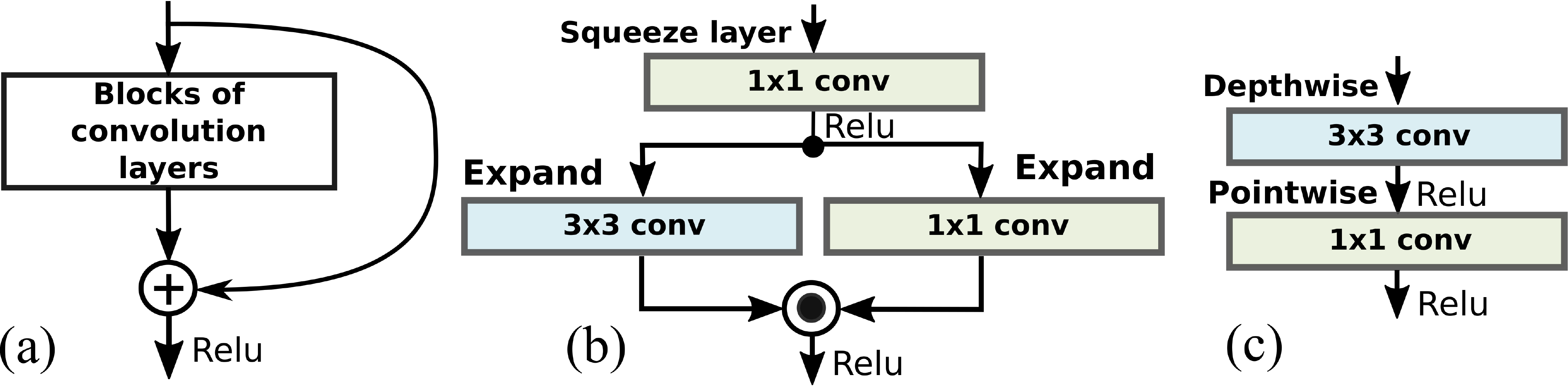}
 \vspace{-0.6cm}
 \caption{Different unit modules used in our architecture designs.
 (a) In the residual module we use the sum operation to combine inputs
 from the shortcut connection and the set of convolution layers;
 (b) fire modules in SqueezeNets combines $1\times 1$ and $3\times 3$
 convolution layers output with a concatenation operation;
 (c) in MobileNets a standard convolution is decomposed into depthwise
 and pointwise convolution layers.
 }
 \label{fig:zoom-in}
\end{figure}

MobileNets \cite{Mobilenets_Howard_2017} are built on
separate filters
factorizing a standard convolution layer into a \emph{depthwise}
and a \emph{pointwise} convolution layers.
A depthwise operation applies a spatial filter to each input channel
(one different filter per channel).
Pointwise filters are the classical $1\times 1$ convolution filter that performs
a linear combination of all channels of the depthwise output.
This factorization has a high impact in the size and the computation
that the model requires.
Figure \ref{fig:zoom-in}(c) illustrates the design we follow to implement
depthwise and pointwise convolution filters.
$3\times 3$ filters are used for the depth wise convolutions.
We add batch normalization and ReLU units after each convolution.

\mypartitle{Feature extraction network.}
%
It is composed of 8 depthwise - pointwise
layers, denoted as Depth/point in Figure~\ref{fig:net-archs}(c).
Pooling mechanisms are included in the form of convolutions with stride 2
(denoted with *).
We also include three residual connections following the design in Figure~\ref{fig:zoom-in}(a).

\mypartitle{Pose regression cascade.}
In each stage, we compose both branches
by 5 depthwise-pointwise modules followed by two $1\times 1$
convolution layers.

%
%

\vspace{-0.1cm}
\subsection{Confidence map ground truth and training}
\label{sec:cpm-training}

We regress confidence maps for the location of the
different body parts and predict vector fields (part affinity fields) for the location 
and orientation of the body limbs.
%
%
The ideal representation of the body part confidence
maps $\PartMaps^{*}$ encodes their locations in the depth image as
Gaussian peaks. Let $\mathbf x_j$ be the ground truth position in the image of
body part $j$. The value $\PartMaps^{*}_j(\mathbf p)$ for pixel $\mathbf p$ in
the target confidence map $\PartMaps^{*}_j$ of the $jth$ part is computed as
\begin{equation}
\PartMaps^{*}_j(\mathbf p) = exp \left(  -\frac{|| \mathbf p_j - \mathbf x_j ||_2^2}{\sigma^2}  \right),
\label{eq:peaks}
\end{equation}
where $\sigma$ is chosen empirically.

Regarding the limbs, the ideal representation $\LimbMaps^{*}$ encodes
with a vector field the
confidence that two body parts are associated, as well as the
information about the orientation of the limbs.
Consider a limb of type $c$ that connects two body parts $j_1$
and $j_2$ (\eg elbow and wrist) and their positions on the depth image are
$\mathbf x_{j_1}$ and $\mathbf x_{j_2}$. The ideal affinity field
at point $\mathbf p$ is defined as
\begin{equation}
\LimbMaps_c^{*}(\mathbf p) =
	\begin{cases}
	\mathbf v & \text{if $\mathbf p$ is on limb $c$}\\
	0         &  \text{otherwise}
	\end{cases},
\end{equation}
where $\mathbf v$ is the unit vector that goes from $\mathbf x_{j_1}$
to $\mathbf x_{j_2}$. The set of pixels that lie on the limb are those
within a distance to the line segment that joins the two body parts.

\mypartitle{Training.}
Supervision is applied at the end of each prediction stage to
prevent the network from vanishing gradients. This supervision is implemented by
two $L_2$ loss functions, one for each of the two branches, between
the predictions $\PartMaps_{\StageIndex}$ and $\LimbMaps_{\StageIndex}$ and the ideal
representations $\PartMaps^{*}$ and $\LimbMaps^{*}$ for stage $s$
%
%
\begin{eqnarray}
\LimbMapsLoss = \sum_{\mathbf p \in \mathbf I} || \PartMaps_{\StageIndex}(\mathbf p) - \PartMaps^{*}(\mathbf p)  ||_2 ^2, \\
\PartMapsLoss = \sum_{\mathbf p \in \mathbf I} || \LimbMaps_{\StageIndex}(\mathbf p) - \LimbMaps^{*}(\mathbf p) ||_2 ^2.
\label{eq:losses}
\end{eqnarray}
The final multi-task loss is computed as:

\begin{equation}
\SkeletonLoss = \sum_{\StageIndex=1}^{\NStages} \left(\PartMapsLoss + \LimbMapsLoss \right),
\label{eq:cpm-loss}
\end{equation}

\noindent where $\NStages$ is the total number of prediction stages.

\vspace{-0.1cm}
\subsection{Body part association}
We use the algorithm presented in \cite{CPMPaf} that uses the part
affinity fields as confidence to associate the different body landmarks
and perform the inference of the 2d pose.
It works in a
greedy fashion exploiting the skeleton tree structure,
analyzing pairs of 
body landmarks that are potentially linked by a limb type and builds
the pose estimate increasingly.

In a nutshell, the method takes as input a given limb type, \eg forearm.
All detected body landmarks that form such limb type (wrists and elbows) are
connected one to one, forming potential limbs.
The connections are weighted by averaging the predicted vector field along the line that
joins the pair of body landmarks.
Connections with high confidence are kept while the other discarded.
Final pose estimates are built by associating the limb candidates (forearm and arm) that
share body landmark candidates (shoulder, elbow and wrist).

\section{Adversarial Domain Adaptation}
\label{sec:domain-adaptation}
 \begin{figure}
 \centering
 \includegraphics[width=\linewidth]{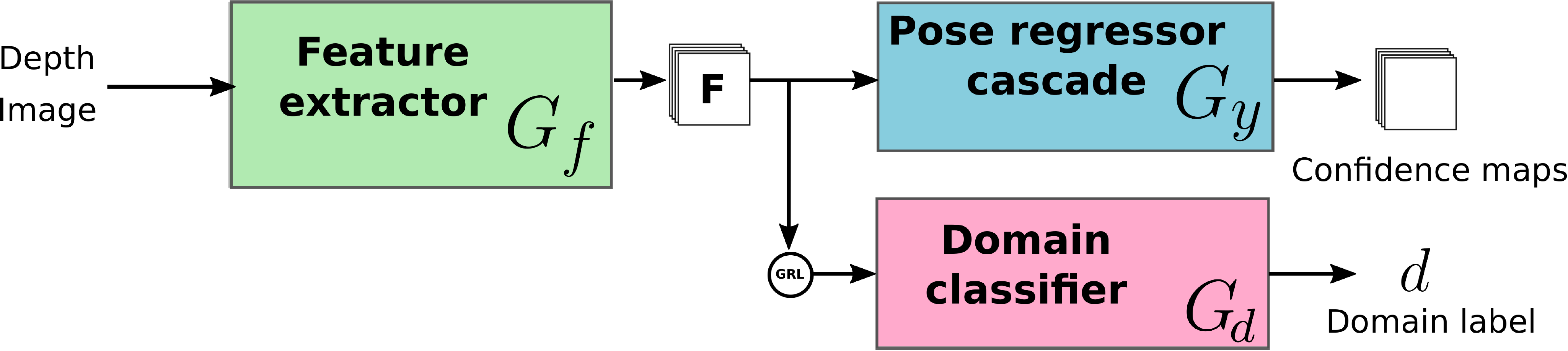}
\vspace{-0.6cm}
 \caption{Architecture used for domain adaptation.
  		 The base architecture is composed of a feature extractor module $\Gf$
		 and a pose regression cascade $\Gy$ (as in Figure~\ref{fig:cpm-sketch}).
		 It is extended with a domain classifier $\Gd$ for depth domain adaptation.}
 \label{fig:res-arch}
 \end{figure}

We are given a source sample set (synthetic depth images)
$\mathcal S=\{(\mathbf{I}_i, \mathbf{x}_i)\}_{i=1}^N$
and a target dataset (real depth images)
$ \mathcal T=\{\mathbf{I}_i\} _{i=1}^M$.
Note that we are only given annotations of 2d keypoint locations
$\mathbf x_i$ for the source samples $\mathbf I_i$. 
The goal is to learn a human pose predictor using the sample $\mathcal S$
which performs well on data from $\mathcal T$, 
by mapping source and target data to an invariant representation.
We follow the approach in~\cite{Ganin_DANN} in which more information can be found.

The distance between the source and target distributions of the input data can be
measured via the H-divergence.
As it is impractical to compute, it can be approximated
by considering the generalization error of a domain classification
problem.
In essence, the distance will be minimum if a
domain classifier is incapable of distinguishing between the samples from the
different domains.
Such domain confusion
can be achieved by learning a mapping from the input data to an 
image representation invariant
across domains.
%

To do so, we rely on the
architecture presented in Figure~\ref{fig:res-arch}.
It is a multi-task architecture comprising three main components.
The first one, $\Gf$ with parameters $\GFParams$ extracts features 
from the input image.
Sharing these features, the branch $\Gy$ with parameters $\GYParams$ 
detects body landmarks and limbs in the image.
In parallel the
branch $\Gd$ with parameters $\GDParams$ classifies the input image 
into a domain label $d\in\{synthetic, real\}$.
%
%
The adversarial adaptation procedure consists of learning a $\Gf$ network 
able to produce high level features sufficient for
body landmark 
detection but which fool the domain classifier $\Gd$.

More formally, let us denote by $\Features_{\mathbf I}$ the internal representation of
image~$\mathbf I$ in the network (features), which is computed as
$\Features_{\mathbf I}=\Gf(\mathbf I; \GFParams)$.
%
We can then define as a measure of domain adaptation the opposite of the
standard cross-entropy loss for the domain classifier
%
\begin{equation}
\DAMeasure(\GFParams, \GDParams) =
-\frac{1}{N + M} \sum_{\mathbf I\in \mathcal T \cup \mathcal S} l_d(\Gd(\Features_{\mathbf I}; \GDParams), I_d),
\label{eq:proxi-distance}
\end{equation}
where $l_d$ is a logistic regression loss and $I_d$ is
the domain label associated with image $\mathbf I$.
$\Gd$ is a
domain classifier such that
$\Gd(\Features_{\mathbf I}; \GDParams) = 1$ if $\mathbf I$ is a real depth image
and 0 otherwise.
%
%
%
Optimizing the domain classifier
is achieved by maximizing
%
\begin{equation}
\DARegularizer = \max_{\GDParams} \DAMeasure (\GFParams,\GDParams).
\label{eq:max-entropy}
\end{equation}
%
%
%
%
%

As shown in \cite{Ganin_DANN}, Eq~(\ref{eq:max-entropy}) approximates the empirical H-divergence,
measuring the similarity of the real and synthetic samples through the 
learned features $\Features_{\mathbf I}$.
Such similarity measure can then be combined with the regression loss to define our adversarial loss:
\begin{equation}
\DALoss(\GFParams, \GYParams, \GDParams) = \SkeletonLoss(\GFParams, \GYParams) + \DomainMixer \max_{\GDParams} \DAMeasure(\GFParams, \GDParams) ,
\label{eq:domain-pose-loss}
\end{equation}
where $\DomainMixer$ represents the trade-off between
landmark localization and domain adaptation, 
and $\SkeletonLoss(\cdot)$ is defined by Eq.~(\ref{eq:cpm-loss}).

The  adversarial learning process then consists of finding the optimal
saddle point of the 'min-max' loss in Eq~(\ref{eq:domain-pose-loss})
by alternating the optimization of the parameters of the body landmark detector
through the minimization
$(\hat{\GFParams},\hat{\GYParams})=\arg\min_{\GFParams, \GYParams} \DALoss(\cdot,\hat{\GDParams})$,
and the maximization
$\hat{\GDParams}=\arg\max_{\GDParams} \DALoss(\hat{\GFParams},\hat{\GYParams},\GDParams)$.
%
%
%
Therefore, $\GFParams$ evolves adversarially to increase the domain
classification confusion while minimizing the error for landmark detection.

In practice we implemented the domain classifier
using a network  composed of two average pooling layers with an intermediate layer of
$1\times 1$ convolution, and  two fully connected
layers before the classification sigmoid function at the end.
We followed  \cite{Ganin_DANN} and included a gradient reversal layer (GRL)
in the architecture to facilitate the joint
optimization of Eq~(\ref{eq:domain-pose-loss}).
The GRL acts as identity function during the forward pass of the
network, but reverses the direction of the gradients from the
domain classifier during backpropagation.
%

%

%
Due to the difference in nature between the pose regression and domain adaptation problems, 
their losess involved in Eq~(\ref{eq:domain-pose-loss}) span different ranges.
Therefore, the trade-off parameter $\DomainMixer$ has to reflect both the importance of the domain classification as well
as to this difference between ranges.
Its setting is detailed in Section~\ref{sec:DomainAdaptation}.

\section{Distilling the pose from a network}
\label{sec:distillation}
\begin{figure}
\centering
\includegraphics[width=1\linewidth]{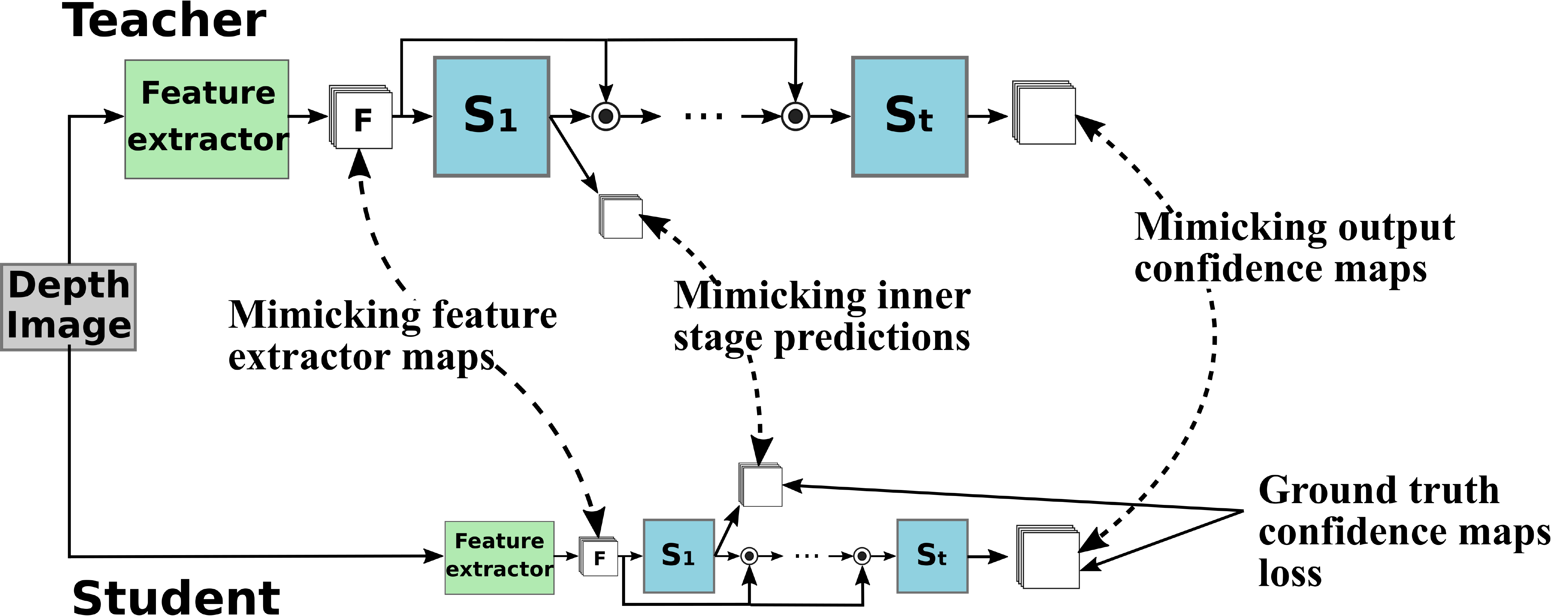}
\caption{Proposed knowledge distillation scheme for body landmark detection
		in a cascade of detectors fashion.
		The student is trained to mimic (in addition to ground truth) the confidence
		map predictions of body landmarks and body limbs from 
		a pre-trained and robust teacher at	the different stages in the cascade 
		of regressors.
		We additionally adopt the learning by hints approach by encouraging the features
		of student to match those of the teacher.
		}
\label{fig:distillation}
\end{figure}
%
Training a CNN model usually relies on data with ground truth.
However, the data samples are often of different complexity levels.
Intuitively, factors like the camera view point, the pose configuration, the  occlusion level (whether due
to  external elements or to its own body), clothing or object artefacts will render the
landmark prediction more or less difficult, an element which is not reflected in the ground truth.
While large over-parametrized networks, when supplied with enough data,
can usually accommodate this diversity during training, smaller more
efficient networks, by attempting to satisfy the ground truth of all data
equivalently, can be more easily trapped in local minima with limited 
generalization.
Several methods have been proposed to improve training.
One of them is curriculum learning \cite{Curriculum_Learning_Bengio},
\ie starting from easier to more complex data samples.
Another direction is knowledge distillation that we now introduce.

The distillation approach can be posed as follows:
given a deep and large teacher network $T_{net}$, we would like to improve the generalization
ability of an efficient student $S_{net}$ using the "knowledge"
acquired by $T_{net}$.
Such knowledge transfer can be of several forms.
First, it usually involves mimicking the output activations of the teacher
as these activations implicitly contains information about the difficulty
of the samples (\eg ambiguity between classes, uncertainty due to difficult imaging conditions).
However, it can also include mimicking some of the hidden layers activation maps.
%
In this case, the activations are referred as \emph{hints} and the goal 
is to drive the student learning towards learning intermediate representations
thought as important from a design process.
%
%
%
%

We learn accurate and efficient models by using a teacher
with high performance
using the distillation strategy illustrated in Figure~\ref{fig:distillation}.
We have investigated several configurations.
First, performing distillation at every stage on the cascade of the
teacher, \ie matching the predictions of the teacher at every prediction stage
in the cascade  to distil the knowledge 
at every prediction stage and to promote 
an early on semantic knowledge distillation.
Note that this contrasts with  the conventional distillation approach which considers 
only the final prediction as knowledge to transfer.
Second, adopting distillation by hints to encourage the student to learn
a data representation similar to that of the teacher.
These two approaches can be combined in an overall pose distillation objective
written as follows:

\begin{equation}
\DistillLoss= \DistillStages + \gamma\DistillHints.
\label{eq:distillation}
\end{equation}

The $\DistillStages$ and $\DistillHints$ losses are described below.

\subsection{Mimicking teacher stage predictions}
%
We couple knowledge distillation with our architecture designs and perform
distillation at the last prediction stages of the cascade.
The motivation is that the teacher's predictions in these last stages contain
valuable information about how the teacher refines predictions,
which may help the student on how to increasingly improve its own predictions.
In practice, we use a weighted sum of losses considering
the teacher's predictions and  the  
ground truth $\SkeletonLoss$ which also introduces information that the 
student should mimic:
\begin{equation}
\DistillStages = (1-\alpha) \TeacherLoss + \alpha \SkeletonLoss,
\label{eq:distillation-stages}
\end{equation}
\noindent where $\SkeletonLoss$ is defined in Eq~(\ref{eq:cpm-loss})
and $\alpha$ is a weighting parameter set to 0.5 in our experiments.
We choose to model $L_{teacher}$ as to match the prediction 
of the $\tau$ ($\tau\geq 1$) last stages of the cascade.
%
%
$L_{teacher}$ is defined as

\begin{equation}
L_{teacher} = \sum_{\StageIndex=0}^{\tau - 1} \left( L^{\PartMaps}_{\NStages-\StageIndex} + 
 L^{\LimbMaps}_{\NStages-\StageIndex}\right),
\label{eq:distillation-output}
\end{equation}

\noindent where $\NStages$ is the number of prediction stages in the teacher's cascade, and
\begin{eqnarray}
L^{\PartMaps}_{\StageIndex} = \sum_{\mathbf p \in \mathbf I} || \PartMaps_{\StageIndex}(\mathbf p) - \PartMaps^{tea}_{\StageIndex}(\mathbf p) ||_2 ^2, \\
L^{\LimbMaps}_{\StageIndex} = \sum_{\mathbf p \in \mathbf I} || \LimbMaps_{\StageIndex}(\mathbf p) - \LimbMaps^{tea}_{\StageIndex}(\mathbf p) ||_2 ^2,
\label{eq:distillation-matchteacher}
\end{eqnarray}
\noindent where $\mathbf p$ is a pixel in image $\mathbf I$,
and $(\PartMaps^{tea}_{\StageIndex}, \LimbMaps^{tea}_{\StageIndex})$ 
are the body part and part affinity fields confidence maps at stage 
$\StageIndex$ of the teacher.
Notice that the number of stages in the teacher and student prediction cascades 
need to be at least $\tau$.

\subsection{Learning with hints}
%
%
%
%
%
We consider this approach by enforcing similarities between the feature extractor outputs
in the teacher and student architectures, as it is a natural choice in our 
architecture designs.
Note that our formulation in Eq~(\ref{eq:distillation-output}) is a form of
distillation with hints for $\tau > 1$.

Thus, denoting by $\mathbf F^{tea}_{\mathbf I}$ the internal representation of 
image $\mathbf I$ of the teacher,
%
we define the hints loss as

\begin{equation}
\DistillHints = ||\mathbf F_{\mathbf I} - \mathbf F^{tea}_{\mathbf I}||_2^2,
\label{eq:distillation-hints}
\end{equation}

%
Note that when performing knowledge distillation with hints it is necessary to tie the dimensions 
of $\mathbf F_{\mathbf I}$ and $\mathbf F^{tea}_{\mathbf I}$.

\section{Experiments and Results}
\label{sec:experiments}
%
In this section we describe the experiments we performed to evaluate
our approaches.
%
%
The experimental protocol focuses on both accuracy and computational
aspects.
We first evaluate the proposed efficient CNN architectures and the impact
of the synthetic dataset for training.
Then we report on the application of the adversarial domain adaptation approach and
its limitations.
%
Finally, we analyze our approach for knowledge distillation and its benefits in our context.


\vspace{-0.2cm}
\subsection{Data}


%
As data, we considered the synthetic and real parts of the publicly available \OurDataset dataset introduced in
Section~\ref{sec:data-domains}, as well as a subset of the CMU-Panoptic dataset, which we briefly describe below.
With both synthetic and real images, we performed data augmentation during training:
rotation by a random angle within  $[-20, 20]$ degrees with a 0.8 probability, 
and image cropping to the $368\times 368$ training size with a probability of 0.9.
Unless stated otherwise, we use DIH-Real as our default (real) dataset in the result section.

\subsubsection{Synthetic data}
\label{sec:synthetic-training}
Train, validation and testing folds contain 230934, 22333 and 11165 synthetic
depth images respectively.
To avoid our pose detector to overfit clean synthetic image details,
we propose to add image perturbation, in particular,
adding real background content which will provide the network with real
sensor noise.

\mypartitle{Adding real background content.}
Obtaining real background depth images (which do not require ground-truth)
is easier than generating synthetic body images.
As background images, we  consider the dataset in~\cite{DepthBackground} containing
1367 real depth images recorded with a Kinect~1 and exhibiting depth
indoor clutter, and
we divided it into training, validation and test folds.
%
In the training case, images were produced on the fly
by randomly selecting one depth image background and body synthetic images,
and generating a depth image using the character silhouette mask.
We simply verified that there was a sufficient depth margin between the
body foreground and the background, adding if necessary an adequate depth constant value to the entire
background image.
While crude, this approach resulted in more  realistic data than the synthetic ones.
Sample results are shown in Fig.~\ref{fig:randomscenario}(e).

\mypartitle{Pixel noise.}
During training we randomly selected 20\% of the body silhouette's
pixels and set their value to zero.

\vspace*{1em}

\subsubsection{Real data}
We consider the following datasets.

\mypartitle{DIH-Real.}
It consists of 16 sequences divided into train, validation and testing
folds comprising 7, 5 and 4 sequences respectively.
We manually annotated a small subset of images for each fold:
1750, 750 and 1000 images within the training, validation, and testing folds, respectively.

\mypartitle{Panoptic.}
We used a subset of the large CMU-Panoptic dataset~\cite{Joo_2017_TPAMI} from
the \emph{Range of Motion (RM)} and \emph{Haggling (H)} scenarios (specifically, RM:171204\underline{ }pose3
and H:170407\underline{ }haggling\underline{ }a3 for training, RM:171204\underline{ }pose5
and H:170407\underline{ }haggling\underline{ }b3 for testing).
To be consistent with our experimental setting, where few labeled data are available,
we randomly considered 2K images  for training and 1K  for testing.

\vspace{-0.2cm}
\subsection{Evaluation metrics}
\mypartitle{Pose estimation performance.}
We use standard precision and recall measures derived from the Percentage
of Correct Keypoints (PCK) evaluation protocol  as performance
metrics~\cite{Yang_PAMI}.
More precisely, after the forward pass of the network, we extract all the
landmark predictions $p$ whose confidence are above a threshold
$\ConfidenceMapThreshold$, and
run the part association algorithm to generate pose estimates from these
predictions\footnote{Note that in this algorithm, landmark keypoints not
associated with any estimates are automatically discarded.}.
Then, for each landmark type, and for each ground truth points $q$,
we associate the closest prediction $p$ (if any) whose distance to $q$
is within a distance threshold $\PCKhDistanceThresh=\PCKhCoef \times \PCKhBBGtHeight$,
where \PCKhBBGtHeight stands for the height of the
bounding box of the person (in the ground truth) to which $q$ belongs to.
Such associated $p$ are then counted as true positives,
whereas the rest of the landmark predictions are counted as false positives.
Ground truth points $q$ with no associated prediction are counted as
false negatives.
Recall and precision values are computed for each landmark type
counting true positives, false positives and false negatives over the dataset.
Finally, the average recall (AR) and average precision (AP) values used to report performance
are computed by averaging the above recall and precision over landmark type and 
over several distance
thresholds \PCKhDistanceThresh by varying $\PCKhCoef \in [0.05,0.15]$.

\mypartitle{Computational performance.}
Model complexity is measured via the number of parameters it comprises,
and the number of frames per second (FPS) it can process when considering
only the forward pass of the network.
This was measured using the median time to process 2K images at resolution
$444 \times 368$ pixels with an Nvidia card GeForce GTX 1050.

\vspace{-0.2cm}
\subsection{Implementation details}

\mypartitle{Image preprocessing.}
The depth images are normalized by scaling linearly the depth values
in the $[0,8]$~meter range into the  $[-0.5,0.5]$ range.

\mypartitle{Network training.}  Pytorch is used in all our experiments.  We
train different network architectures with stochastic gradient descent with the
momentum set to 0.9, the decay constant to ${5 \times 10^{-4}}$, and the batch size to~10.
%
We uniformly sample values in the range $[4 \times 10^{-10}, 4 \times 10^{-5}]$
as starting learning rate and decrease it by a factor of 10 when the validation
loss has settled.
All networks are trained from scratch and progressively, \ie to train
network architectures with $\StageIndex$ stages, we initialize the network with the
parameters of the trained network with ${\StageIndex-1}$ stages.
Unless otherwise stated, each model was trained with synthetic data for
13 epochs and finetuned with real images for 100 epochs.

\mypartitle{Tested models and notation.}
%
%
Residual, mobile and squeeze pose machines are referred to as RPM, MPM and SPM
respectively.
We set $N_w=64$ (number of features) as the default value and we specify when it changes.
We add a postfix to specify the number of stages that a model comprises.
For example, RPM-2S is the residual pose machine configuration with 2
prediction stages in the cascade of predictors.

\vspace{-0.2cm}
\subsection{Performance-efficiency trade-off}
%
\begin{table*}[t]
\centering

\begin{tabular}{l | c c c c | c c c c c c}

\Xhline{2\arrayrulewidth}

\multicolumn{11}{c}{DIH} \\

\Xhline{2\arrayrulewidth}

&\multicolumn{4}{|c|}{} & \multicolumn{3}{c}{All body} & \multicolumn{3}{c}{Upper body} \\

\cline{6-11}

Architecture &  \# Stages  & \# Features & \# Parameters & FPS  & AP  & AR & F-Score &  AP  & AR & F-Score \\

\Xhline{2\arrayrulewidth}
\rowcolor{Gray}
HG \cite{HourGlass}& 2 & - & 12.93 M & 8.7 & 84.62 & 86.26 & 0.85 & 86.45 & 93.70 &  0.90  \\

\Xhline{2\arrayrulewidth}

CPM-1S \cite{CPMPaf} & 1 & 128 & 8.38 M & 18.6 & 92.10 &  86.44 & 0.89 & 96.12 & 94.99 & 0.95 \\
\rowcolor{Gray}
CPM-2S \cite{CPMPaf} & 2 & 128 & 17.07 M & 11.2 & 94.03 & 88.96 & \textbf{0.91} & 96.36 & 94.98 & \textbf{0.96} \\

\Xhline{2\arrayrulewidth}

RPM-1S & 1 & 64 & 0.51 M & 56.7 & 90.86 & 72.07 & 0.80 & 94.28 & 87.77 & 0.91 \\
\rowcolor{Gray}
RPM-2S & 2 & 64  & 2.84 M & 35.2 & 94.84 & 86.41 & 0.90 & 96.39 & 93.91 & 0.95  \\

RPM-3S & 3 & 64  & 5.17 M & 20.8 & 93.96 & 87.72 & \textbf{0.91} & 97.26 & 94.72 & \textbf{0.97} \\

\Xhline{2\arrayrulewidth}
\rowcolor{Gray}
RPM-2S & 2 & 128 & 10.5 M  & 12.5 & 93.52 & 86.07 & 0.90 & 95.95 & 94.02 & 0.95 \\

\Xhline{2\arrayrulewidth}
MPM-1S & 1 & 64  & 99.8 K   & 134.4 & 88.52 & 71.36 & 0.79 & 92.88 & 84.74 & 0.89  \\

\rowcolor{Gray}
MPM-2S & 2 & 64  & 168.2 K  & 112.3 & 92.56 & 78.63 & 0.85 & 94.97 & 89.92 & 0.92 \\

MPM-3S & 3 & 64  & 236.5 K  & 95.8  & 92.40 & 82.79 & 0.87 & 95.23 & 92.36 & 0.94 \\

\rowcolor{Gray}
MPM-4S & 4 & 64  & 304.9 K  & 84.3  & 91.27 & 84.06 & \textbf{0.88} & 95.27 & 91.87 & \textbf{0.94} \\

\Xhline{2\arrayrulewidth}
SPM-1S & 1 & 64  & 308.8 K & 70.6 & 89.64 & 61.58 & 0.73 & 93.18 & 71.38 & 0.81 \\

\rowcolor{Gray}
SPM-2S & 2 & 64  & 455.9 K & 60.6 & 91.78 & 81.68 & 0.86 & 95.82 & 91.47 & 0.94 \\

SPM-3S & 3 & 64  & 660.0 K & 53.3 & 92.63 & 81.98 & \textbf{0.87} & 96.43 & 90.84 & \textbf{0.94} \\

\rowcolor{Gray}
SPM-4S & 4 & 64  & 921.0 K & 47.2 & 93.13 & 81.36 & 0.87 & 96.30 & 90.58 & 0.93\\

\Xhline{2\arrayrulewidth}

\multicolumn{11}{c}{Panoptic} \\

\Xhline{2\arrayrulewidth}

& \multicolumn{4}{|c|}{} & \multicolumn{3}{c}{All body} & \multicolumn{3}{c}{Upper body} \\

\cline{6-11}

Architecture &  \# Stages  & \# Features & \# Parameters & FPS  & AP  & AR & F-Score &  AP  & AR & F-Score \\

\Xhline{2\arrayrulewidth}
\rowcolor{Gray}
HG     \cite{HourGlass} & 2 & -   & 12.93 M  & 8.7   & 85.14 & 91.06 & 0.88 & 86.0  & 91.0  & 0.88  \\

CPM-2S \cite{CPMPaf}    & 2 & 128 & 17.07 M  & 11.2  & 96.73 & 91.66 & \textbf{0.94} & 96.75 & 92.07 & \textbf{0.94} \\

\rowcolor{Gray}
RPM-3S 					& 3 & 64  & 5.17 M   & 20.8  & 97.42 & 91.48 & \textbf{0.94} & 97.43 & 91.48 & \textbf{0.94} \\

MPM-4S 				    & 4 & 64  & 304.9 K  & 84.3  & 96.43 & 89.17 & 0.92 & 97.0  & 89.06 & 0.93 \\

\rowcolor{Gray}
SPM-3S 					& 3 & 64  & 660.0 K  & 53.3  & 95.44 & 89.24 & 0.92 & 96.30 & 90.35 & 0.93 \\

\Xhline{2\arrayrulewidth}
\end{tabular}

\vspace*{-1mm}

\caption{Performance (\%) on the test set of real depth images and architecture components for the different tested
	     pose machines instances.
	     Upper body comprises only head, neck, shoulders, elbows and wrists.
}

\vspace*{-2mm}

\label{tab:speedvsacc}
\end{table*}

Table~\ref{tab:speedvsacc} compares the performance of our proposed
architecture configurations.
%
%
We report both the average recall and average precision for all landmark
types in the skeleton model and for the upper body, \ie
\emph{head, neck, shoulders, elbows and wrists} since upper body detection might be
sufficient for any given applications (\eg HRI).
We compare the models' performance with the F-Score metric (harmonic mean of recall and precision).
The table also compares the FPS and the number of trainable parameters of the
different architectures.

\begin{figure}
\centering
\includegraphics[width=0.49\linewidth]{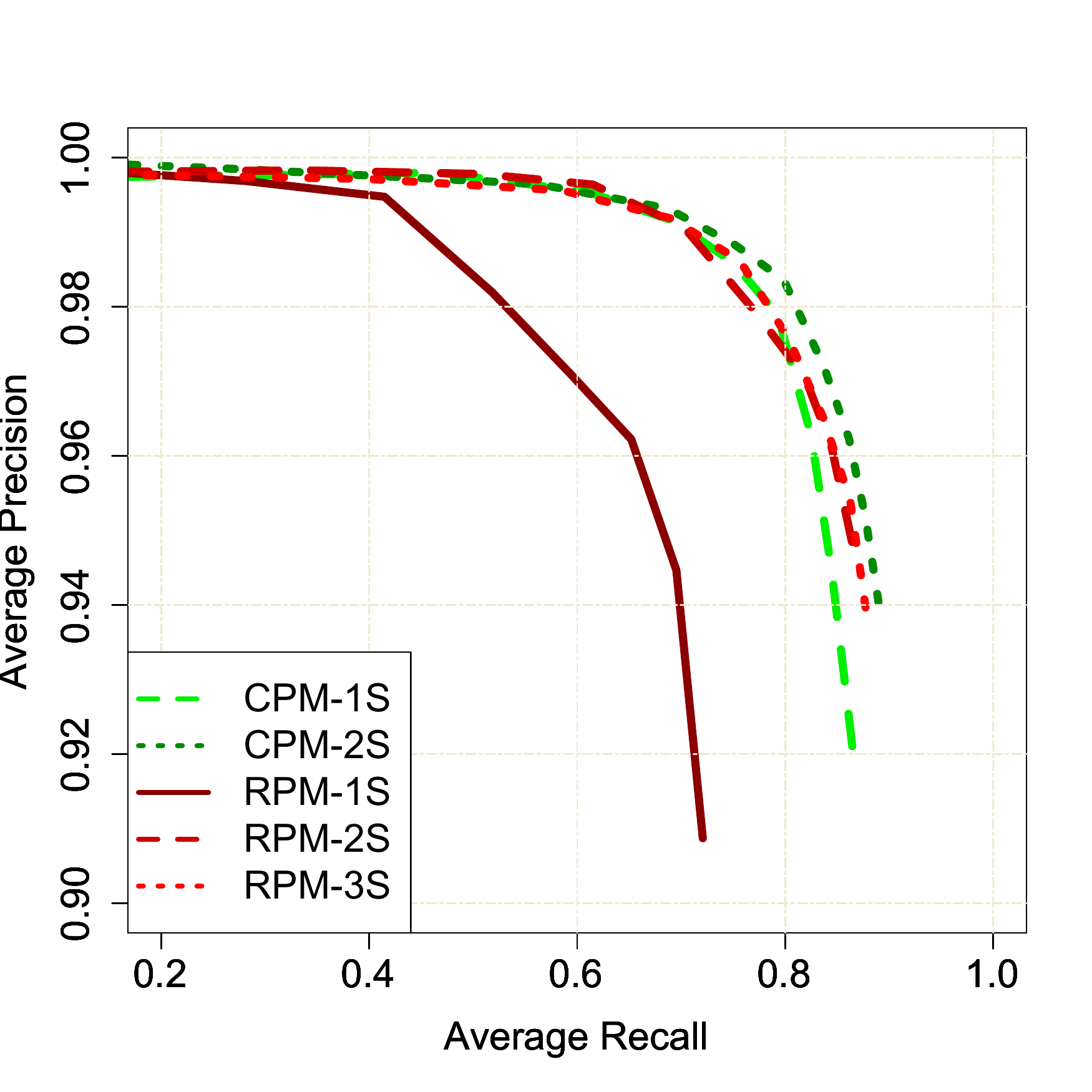}
\includegraphics[width=0.49\linewidth]{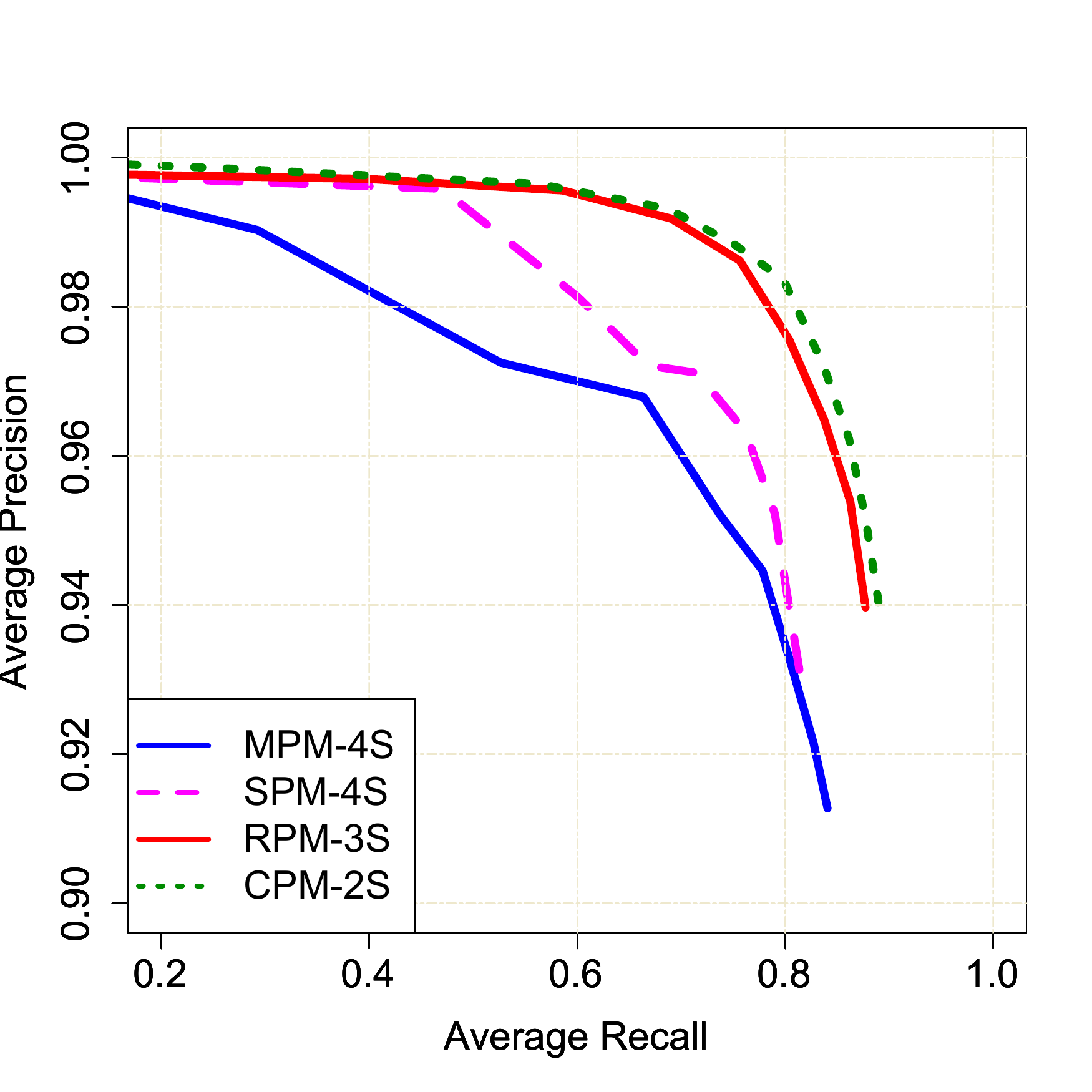}
\vspace*{-0.5cm}
\caption{Precision-recall curves obtained by varying the landmark detection threshold 
		$\ConfidenceMapThreshold$.
		Left: comparison between the baseline CPM and the  RPM model with
		different number of prediction stages.
		%
		%
		Right: performance comparison of our  pose machines instances in their
		deepest version.
}
\vspace*{-2mm}
\label{fig:rpm-stages}
\end{figure}

\begin{figure}[tb]
\includegraphics[width=.49\linewidth]{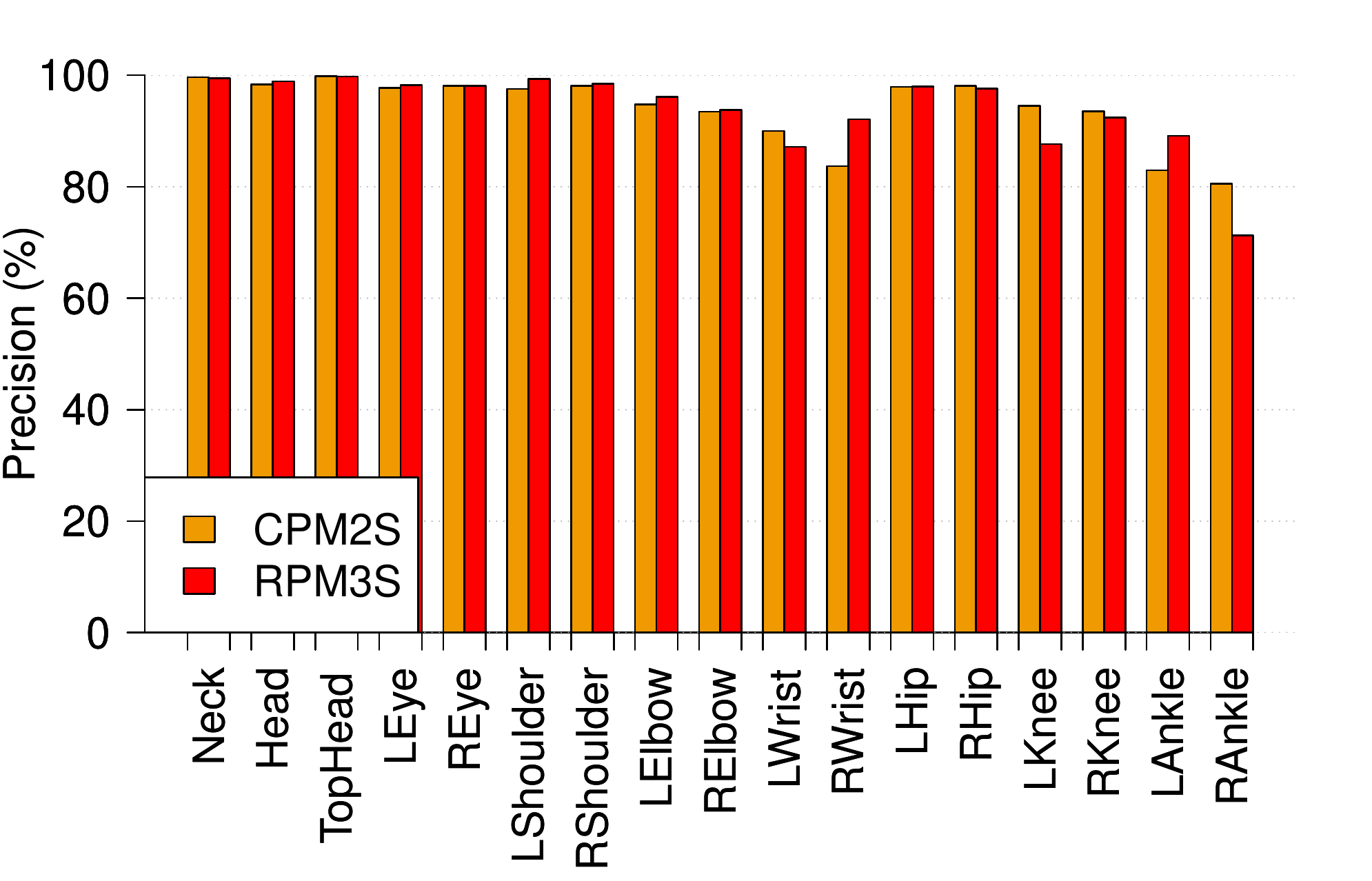}
\includegraphics[width=.49\linewidth]{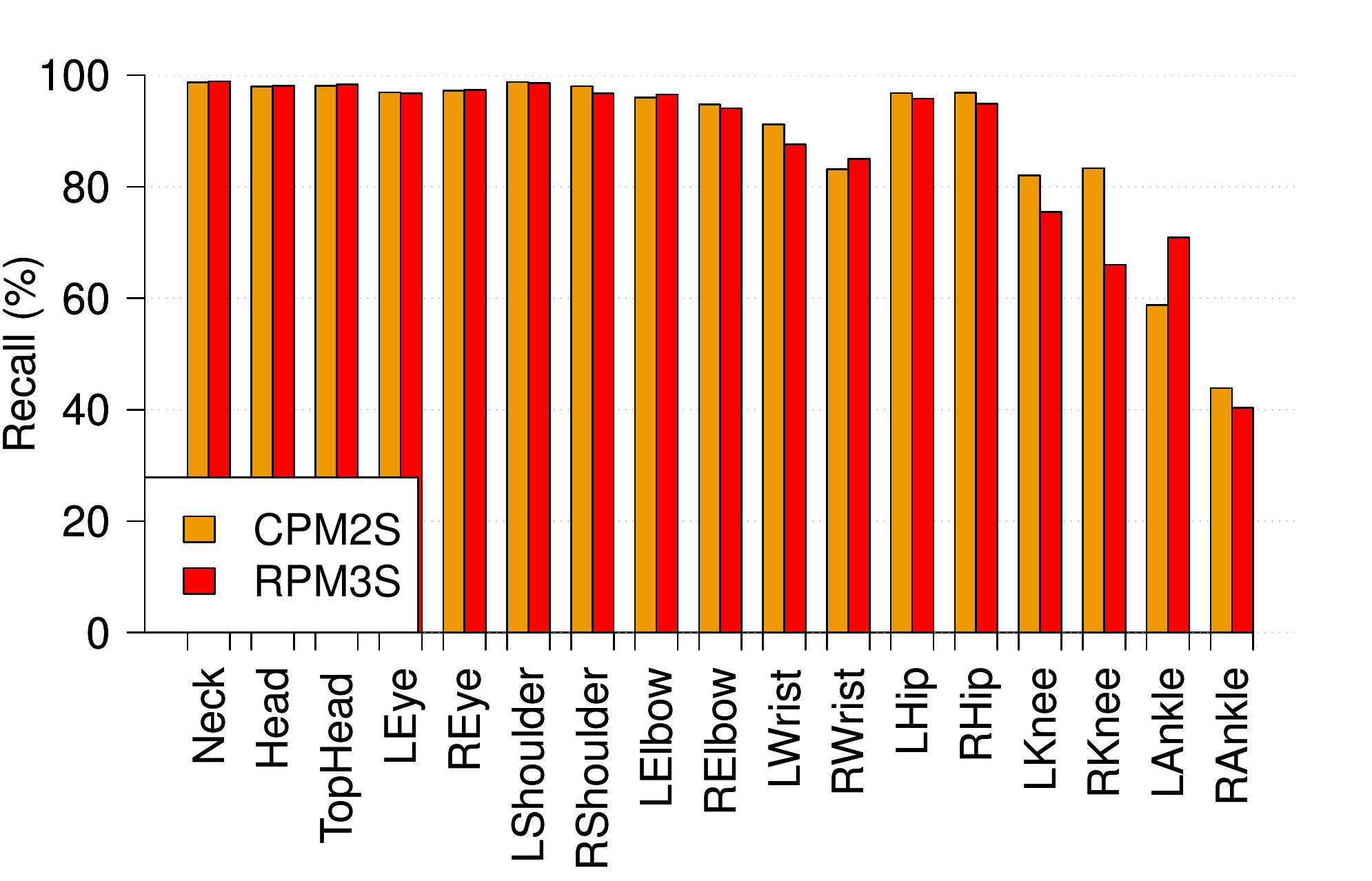}
\vspace{-0.25cm}
\caption{Precision (left) and recall (right) per body landmark for the RPM-3S and the baseline CPM-2S.}
\label{fig:bar-plots}
\vspace*{-2mm}
\end{figure}

\begin{figure*}
\centering
\includegraphics[width=0.8\linewidth]{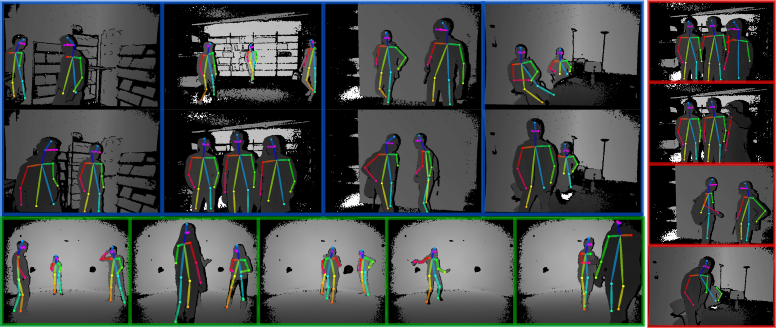}
\vspace{-1mm}
\caption{Left: Pose estimation examples from the RPM-3S landmark detector
		on the DIH dataset (blue squares) and CMU-Panoptic dataset (green squares).
		Right: Examples of failures cases. They often occur for specific pose or
		clothings or under occlusion of people or objects (\eg bags).
}
\vspace*{-2.5mm}
\label{fig:good-examples}
\end{figure*}

\subsubsection{Comparison with the Convolutional Pose Machine (CPM) baseline}

%
%
We consider the CPM architecture presented in~\cite{CPMPaf} as the main baseline.
%
%
As in the original work, the architecture parameters in the feature
extractor module were initialized using the first 10 layers of the
VGG-19 network.
Parameters in the cascade of detectors were trained from scratch.
To accommodate the need for the 3 channel image input expected by VGG-19,
the single depth channel is repeated three times.
We report the results for architectures comprising up to two stages in the prediction cascade since no substantial performance increase was obtained after the second~stage.

We train RPM network architectures with 1, 2 and 3 stages.
Figure~\ref{fig:rpm-stages}(a) shows the precision-recall curves over the
testing set of real depth images, obtained 
%
by varying the landmark detection threshold \ConfidenceMapThreshold.
%
%
We summarize these curves in Table~\ref{tab:speedvsacc} taking the performance
with the highest F-Score.
%
As can be seen,  our RPM models perform as well
as the baseline but is lighter and faster.
This is specially the case for RPM-2S that shows similar performance as CPM-2S
but comprises 6 times less parameters and is 3.14 times faster.
Interestingly, we can notice from Figure~\ref{fig:rpm-stages}(a) that the smaller
complexity of the feature extractor of RPM-1S leads to degraded
performance compared to the baseline CPM-1S.
%
This gap is filled once context is introduced with the second stage (see RPM-2S and CPM-2S curves).
Adding an extra stage (RPM-3S) slightly improves
the model performance while still being faster than the baseline.
In particular for the upper body parts where it now slightly
outperforms the baseline.
Figure~\ref{fig:bar-plots} compares the per body landmark precision and accuracy
for RPM-3S and CPM-2S models, 
where we can notice very high performance achieved for the upper body landmarks.

\mypartitle{Number of feature channels.}
%
%
We set $N_w=128$ and train the RPM models.
We report the results for RPM-2S with this configuration in Table~\ref{tab:speedvsacc}.
%
%
Note how the results of the RPM-2S with $N_w=64$ and RPM-2S with $N_w=128$ are very similar
showing that more feature channels does not bring more benefits.
Given its higher computational complexity setting
$N_w=64$ is a good accuracy-speed trade-off.

\mypartitle{Qualitative analysis.}
Figure~\ref{fig:good-examples} shows examples of the pose estimation
algorithm using the body landmarks and limbs output of our RPM-2S model.
%
%
Note that our model is capable of producing strong confidence maps that produce
accurate estimates even for the eyes and in the presence of self and person occlusions, profile views,
and different body pose configurations and silhouette shapes.
The main challenges for our model include strong changes in the person silhouette (backpacks, big jackets)
and person proximity and occlusions as illustrated by failure cases in Figure~\ref{fig:good-examples}(b).

\subsubsection{Efficient pose machines}
%
%
%
The performance of the MPM and SPM models are presented in Table~\ref{tab:speedvsacc}.
We can first notice that they are more efficient in terms of FPS and number
of trainable parameters than the  CPM and RPM models.
For example, MPM-4S contains 55.9 times less 
parameters than CPM-2S model and is 7.5 times faster with only a decrease of 0.03
in F-Score.
SPM-4S, on the other hand is 4.2 times faster than CPM-2S with a decrease of
0.04 in F-Score.

We note that increasing the number of prediction stages improves the models' F-Score.
As with the RPM models, we observe that the biggest improvement appears when introducing a second stage.
The additional stages help refining prediction,
for instance by greatly improving the landmark detection recall
while the precision saturates or slightly degrades
(compare the MPM-2S and MPM-4S results),
but in general 
the results often start saturating after the 3rd stage.

Figure~\ref{fig:rpm-stages}(b) illustrates the precision-recall curves for
the deepest version of our efficient models.
Among the architectures we investigate, the fastest are the MPM designs,
followed by SPM models.
Nevertheless, the best speed-accuracy trade-off are given by the RPM-2S model
when the focus is on accuracy, and MPM-4S when it is on speed.

\subsubsection{Comparison with state-of-the-art methods}
\label{sec:hg-baseline}
In addition to the CPM baseline, we compared our methods
with the stacked Hourglass framework~\cite{HourGlass}.
We used the same network architecture and training protocol (initialization, learning rate, optimizer) 
proposed by the authors.
For a fair comparison we followed our protocol regarding the data, 
trained the model with synthetic images for 13 epochs and finetuned it
with real data for 100 epochs.

Results in Table~\ref{tab:speedvsacc} show that
our efficient architectures outperform this baseline in  efficiency and accuracy, on both the DIH and Panoptic datasets. 
For example, MPM-4S is 9.6 times faster, 42.2 times smaller, with an F-Score of 0.88 (on DIH)
compared to 0.85 for the Hourglass network.

\subsubsection{Experiments with CMU-Panoptic dataset}
\label{sec:panoptic-data}
The results on this dataset are shown in Table I (bottom part). 
We report only the results obtained by the best performing models on the DIH-Real dataset.
Our proposed RPM-3S outperforms the Hourglass baseline.
It also performs similarly than the CPM baseline.
The proposed efficient architectures MPM-4S and SPM-3S outperform the Hourglass baseline by a margin 
of 0.04 in the F-Score and follow closely the RPM-3S and baseline CPM-2S with a difference of 0.02 in the F-Score.

\vspace{-0.2cm}
\subsection{Training with synthetic data analysis}
%
%
\begin{figure}
  \vspace{-2mm}
\centering
\includegraphics[width=0.49\linewidth]{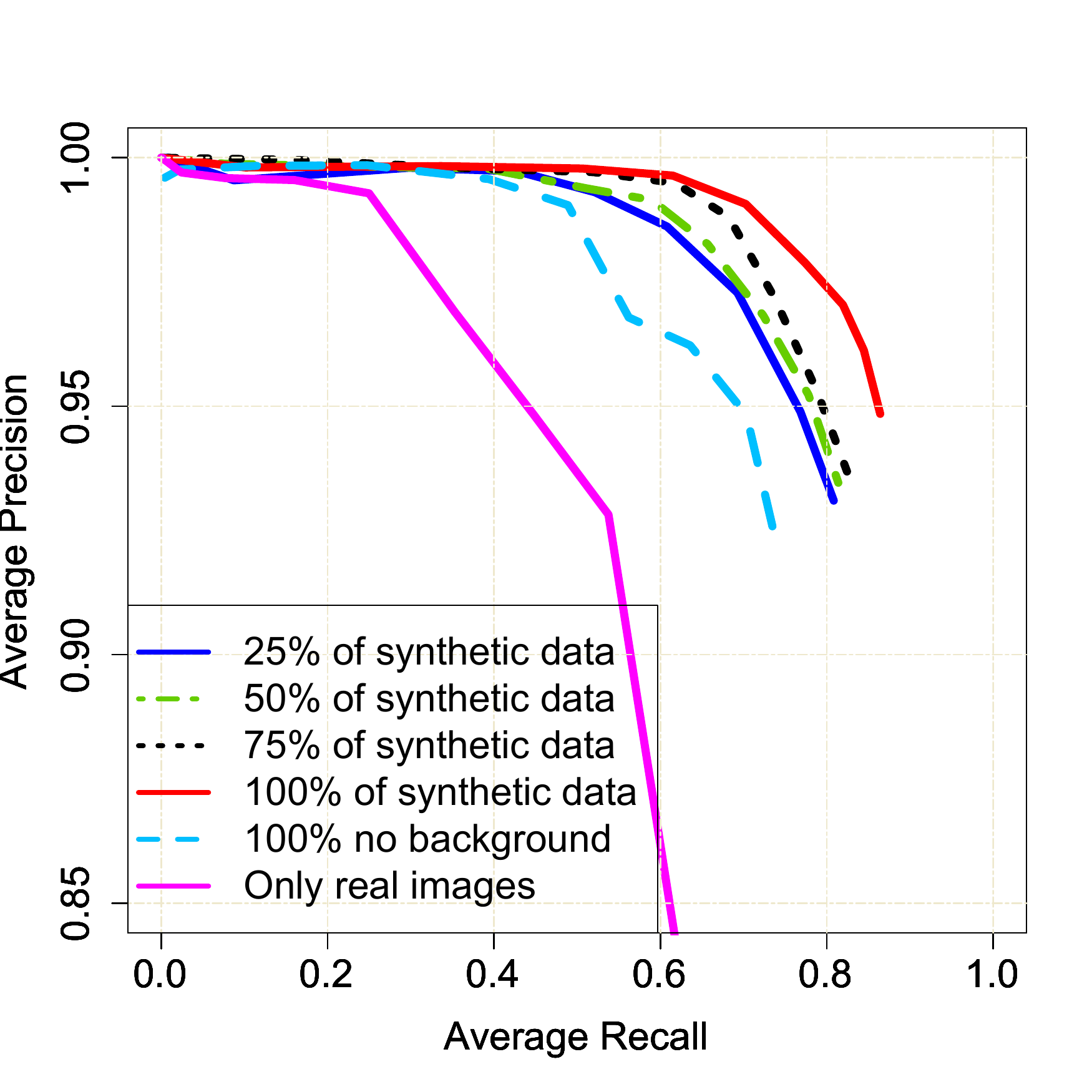}
\includegraphics[width=.49\linewidth]{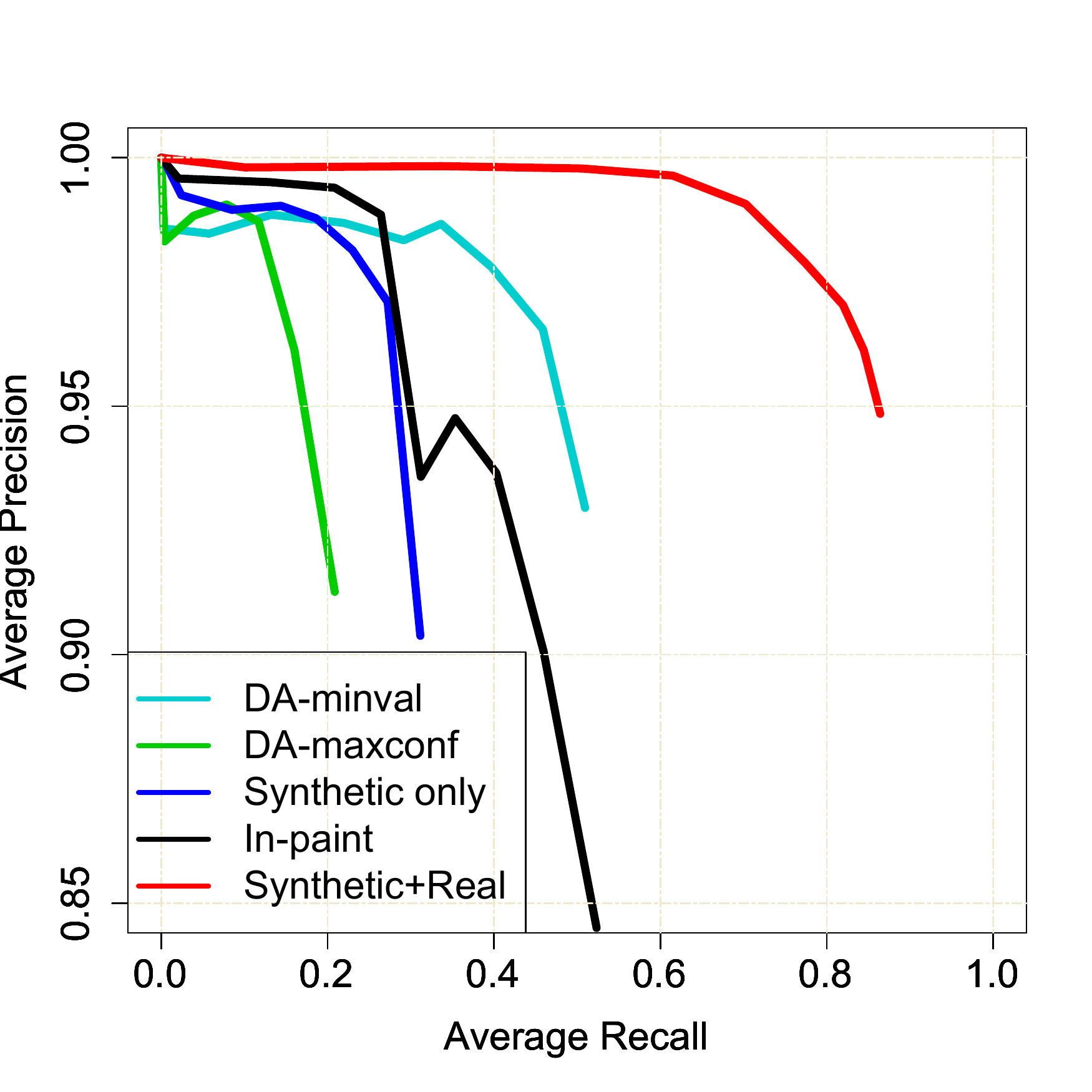}
\vspace{-0.7cm}
\caption{Left: evaluation of the use of synthetic data for learning robust pose estimation models.
		Right: comparison of different techniques for synthetic-to-real domain adaptation.
 }
\label{fig:synthetic-training-da}
\vspace{-2mm}
\end{figure}

We validate the use of the synthetic data to learn accurate models.
%
To this end, we randomly split the synthetic training data in partitions comprising $25\%$,
$50\%$, $75\%$ and $100\%$ of it.
These partitions contain images with one and two people.
We train RPM-2S models with the different synthetic data partitions and then finetune
the result with real data.
We use the same quantity of labeled data to finetune in all cases (1750 images).
Unless otherwise stated, during training we consider all image transformations
for data augmentation (background fusion, pixel drop, cropping and rotation).
Table~\ref{tab:dataset-folds} shows the obtained average recall and precision
%
on real data, before and after finetuning. The following conclusions can be made.

\begin{table}
\centering
\begin{tabular}{c | c c c | c c c}
\Xhline{2\arrayrulewidth}
  & \multicolumn{3}{c}{Synthetic only} & \multicolumn{3}{c}{Synthetic + Finetuning} \\

\cline{2-7}

Synthetic data $\%$ & AP   & AR   & F-Score & AP     & AR    & F-Score \\

\Xhline{2\arrayrulewidth}
\rowcolor{Gray}
25 $\%$              & 94.19 & 26.28 & 0.41    & 93.09 & 80.83 & 0.86 \\

50 $\%$              & 95.10 & 29.51 & 0.45    & 93.45 & 81.38 & 0.87 \\
\rowcolor{Gray}
75 $\%$              & 91.68 & 30.14 & 0.45    & 93.68 & 82.42 & 0.88 \\

100 $\%$             & 90.37 & 31.14 & 0.46    & \textbf{93.96} & \textbf{87.72} & \textbf{0.91} \\

\rowcolor{Gray}

100 $\%$ no BG       & 80.67 & 3.35 & 0.06    & 92.58 & 73.46 & 0.82 \\

\Xhline{2\arrayrulewidth}

\end{tabular}

\vspace*{-1mm}

\caption{Comparison of performance obtained with RPM-2S when trained with different synthetic data partitions.
}
\label{tab:dataset-folds}


\end{table}

\subsubsection{Amount of synthetic data}
%
%
%
%
Performance increases with more synthetic data, both
%
before and after finetuning.
Naturally, the visual features mismatch between the synthetic and real data provokes
low performance when training only with synthetic data.
Nevertheless the gap is covered once finetuning on real data is applied,
particularly regarding the recall.
Figure~\ref{fig:synthetic-training-da}(a) shows the average precision-recall
curves training with the different data partitions and applying finetuning.

\subsubsection{Adding realism to synthetic images}
We validate our strategy of fusing real background with synthetic depth images
to prevent overfiting.
To this end, we trained the RPM-2S models with $100\%$ of the synthetic data, helding out
the background fusion transformation, and then applied finetuning.
We report the results in Table~\ref{tab:dataset-folds}.
Observe that without the addition of real background, our model overfits to the synthetic data
details and performs poorly on real data.
Interestingly, note as well that even after finetuning on real data the performance is not entirely recovered.
%
The model even performs lower than using only $25\%$ of the synthetic data.
Intuitively, fusing background with synthetic images work as a regularizer that
prevents overfitting to synthetic image details.

\subsubsection{Multiple people data}
\begin{table}
\vspace{-1mm}
\centering
\begin{tabular}{c | c c c c c c}
\Xhline{2\arrayrulewidth}

  & \multicolumn{3}{c}{All test data} & \multicolumn{3}{c}{Data with person occlusions} \\
\cline{2-7}

Data fold      & AP    & AR    & F-Score  & AP    & AR    & F-Score \\
\Xhline{2\arrayrulewidth}

\rowcolor{Gray}
1P-Fold  & 92.51 & 80.22 & 0.86 & 91.10 & 82.92 & 0.87\\

2P-Fold & 92.43 & 82.44 &\textbf{0.87} & 93.33 & 84.81 & \textbf{0.89} \\

\Xhline{2\arrayrulewidth}

\end{tabular}

\vspace*{-1mm}

\caption{Performance obtained with the RPM-2S model when combining synthetic
		 data with only single person images (1P-Fold) and 2 people images (2P-Fold)
		 for training.
}
\label{tab:dataset-two-people}
\end{table}
We study the importance of having synthetic training images with two people
before finetuning with real data.
To this end, we define two folds of 100K images for training.
The first one (1P-Fold) contains only images with  one person;
the second one (2p-Fold) contains 50K images with one person and 50K images with
two people.
The resulting performance is reported in Table~\ref{tab:dataset-two-people}
where we also provide results on a test subset where people are very close
or occlude each other (see Figure\ref{fig:good-examples}(b)).
The subset contains 211 images with three people where their ground truth
bounding boxes overlap between 12.38\% and 15.4\%.
We note that using images with two people helps generalization.

\subsubsection{Training with real data only}
We train the RPM-2S model only with our small real depth image annotated sample.
Figure~\ref{fig:synthetic-training-da}(a) shows the performance curve.
Our real dataset sample is not large enough to prevent our model from overfiting
and performs worse than using the synthetic data without background fusion.

\vspace{-0.2cm}
\subsection{Domain adaptation}
\label{sec:DomainAdaptation}

\begin{table}
\vspace{-1mm}

\centering
\begin{tabular}{l | c c c c}
\Xhline{2\arrayrulewidth}

Adaptation method & CR & AP & AR & F-Score \\

\Xhline{2\arrayrulewidth}

ADA - Min Loss  & 0.20 & \textbf{92.95} & \textbf{50.94}  &\textbf{0.66} \\

\rowcolor{Gray}
ADA - Max Confusion  & 0.50 & 91.26 & 20.86  & 0.40  \\

\Xhline{2\arrayrulewidth}

Only synthetic          & 0.04* & 90.37 & 31.14 & 0.46   \\

\Xhline{2\arrayrulewidth}

\rowcolor{Gray}
In-paint preprocessing  & 0.03* & 84.49 & 52.33 & 0.65    \\ 

Finetunning             & 0.15* & \textbf{93.96} & \textbf{87.72} & \textbf{0.91} \\

\Xhline{2\arrayrulewidth}

\end{tabular}
\vspace{-1mm}

\caption{Comparison of performance obtained with the different techniques for
		domain adaptation using the RPM-2S model.
		CR stands for the confusion rate obtained in the domain classification task.
		Values with * were obtained by training a domain classifier using the
		corresponding model feature extractor as fixed features.
		In-paint preprocessing corresponds to the application of in-painting
		before applying the detector.
}
\vspace{-2mm}
\label{tab:domain-adaptation}
\end{table}
In these experiments we use the unlabeled training set of real images
comprising 4828 Kinect~2 depth images as the target domain.
We first train the feature extractor
and cascade of predictors of the RPM-2S model with synthetic data for $200K$ iterations.
Adversarial domain adaptation (ADA) is then performed by jointly training the domain
classifier, feature extractor and cascade of predictors for another $100K$
iterations\footnote{Note that since ADA requires no annotation, it could be possible to  collect and use
  much more images than the 4828 images, with the expectation of obtaining better results.
  This was not done here and left as future work.}.
Following common practices
we gradually updated the trade-off parameter $\lambda$ of
eq~(\ref{eq:domain-pose-loss}) according to the training progress as
$\lambda_p = \frac{2 \Lambda}{1+ exp(-10p) } - \Lambda$,
where $p=t/T$, with $t$  the current iteration and $T=100K$. 
We set  $\Lambda=100$  so that the two
losses in Eq~(\ref{eq:domain-pose-loss}) are in the same range.

\subsubsection{Adaptation criteria}
During adaptation we monitored validation losses for body landmark detection and 
domain classification.
After $T$ iterations we selected domain adapted models according 
to two criteria: 1) the minimum of a pose validation loss,
and 2) maximum confusion (sum of false positives and 
false negatives for the domain classification task) on a validation set.
Table~\ref{tab:domain-adaptation} reports the  results
where we also include the performance of models trained without adaptation.
The adversarial domain adaptation (ADA) framework aims at learning invariant  features
across domains.
Nevertheless, we note that when maximum confusion is achieved the body landmark detection task is greatly hampered,
and the model performs even worse
than non-adapted models.
In contrast, the model selected using a validation error on body landmark detection
outperforms the non-adapted models while still achieving some level of domain
confusion.
We can notice that the recall is greatly affected.
Certainly, a major difference between synthetic and real images is the lack of data
around external edges that form the limbs extremities and silhouette.
These are also the places where pose information is available.
Confusing the domain classifier means that features exploiting this lack of data are removed, which
hurts the  body landmark detection performance.

\subsubsection{Inpainting and finetuning  methods.}
To understand the limits of our unsupervised ADA approach, we compare it
with the performance of inpainting and finetuned models.
Figure~\ref{fig:synthetic-training-da}(c) shows the precision-recall curves of the different methods,
while Table~\ref{tab:domain-adaptation} summarizes these curves with the maximum F-Score obtained.
We see that ADA slightly outperforms the in-painting approach.
Indeed, the latter aims to fill the missing depth information as an
image preprocessing step for non-adapted models.
Observe however that in-painting greatly reduces precision, introducing
artifacts in the image that are later confused as body landmarks or limbs.
On the other hand, finetuning directly the network initially training on synthetic images  with even
a small amount of labeled data  greatly improves its generalization capabilities.
\begin{table}
  \vspace{-2mm}
\centering
\begin{tabular}{c | c | c | c c c }
\Xhline{2\arrayrulewidth}

\multicolumn{6}{c}{DIH} \\

\Xhline{2\arrayrulewidth}

Student & Teacher & Distil type & AP & AR & F-Score \\
\Xhline{2\arrayrulewidth}
\rowcolor{Gray}

CPM-2S	& - & - & 94.03 & 88.96 & 0.91 \\
\Xhline{2\arrayrulewidth}
MPM-2S	& - & - & 92.56 & 78.63 & 0.85 \\
\rowcolor{Gray}
MPM-4S	& - & - & 91.27 & 84.06 & 0.88 \\

\Xhline{2\arrayrulewidth}
\Xhline{2\arrayrulewidth}

MPM-1S	& CPM-2S & Stagewise & 88.55  & 71.96 & 0.79\\
\rowcolor{Gray}
MPM-1S	& CPM-2S & Stagewise + Hints & 89.89 & 76.61 & 0.83 \\

\Xhline{2\arrayrulewidth}

MPM-2S  & CPM-2S & Standard ($\tau=1$)& 90.98 & 80.60  &  0.85 \\
\rowcolor{Gray}
MPM-2S	& CPM-2S & Stagewise & \textbf{92.10} & \textbf{83.98} & \textbf{0.88} \\

MPM-2S  & CPM-2S & Stagewise* ($\tau=2$) & 92.42 & 81.60  &  0.87 \\
\rowcolor{Gray}
MPM-2S	& CPM-2S & Stagewise + Hints & 90.91 & 80.19 & 0.85 \\

\Xhline{2\arrayrulewidth}

\multicolumn{6}{c}{Panoptic} \\

\Xhline{2\arrayrulewidth}

\rowcolor{Gray}
CPM-2S           & -     & -          & 96.73 & 91.66 & 0.94 \\

MPM-2S 			 & -     & -          & 96.25 & 86.66 & 0.91 \\
\rowcolor{Gray}
MPM-4S 		     & -     &  -         & 96.43 & 89.17 & 0.92 \\

\Xhline{2\arrayrulewidth}
\Xhline{2\arrayrulewidth}

MPM-2S 		     & CPM-2S& Stagewise & 95.08 & 90.33 & 0.92 \\

\Xhline{2\arrayrulewidth}

\end{tabular}
\vspace{-1mm}
\caption{Knowledge distillation experiments.
		%
		%
		Methods with - in the 'Teacher' and 'Distil type' fields 
		indicate that the model has been trained
		without distillation 
		and their values were taken from 
		Table~\ref{tab:speedvsacc} (\ie these are the baseline results).
}
\label{tab:distillation-results}
\vspace{-2mm}
\end{table}
%
%

\vspace{-0.2cm}
\subsection{Pose knowledge distillation}

We perform knowledge distillation using CPM-2S as the teacher.
We select MPM-1S and MPM-2S models as students given that the MPM design
is the most efficient of our pose machines.
Our students are trained from scratch.
We performed distillation first with synthetic data during 10 epochs,
then with real images for 100 epochs.
The learning rate is updated in a linear fashion and we set $\gamma$ to 1.
Table~\ref{tab:distillation-results} reports the results.
Our proposed approach of performing distillation at the last  stages in cascade
is referred to as \emph{stagewise}.
We also experiment by matching only the final activation maps of the teacher
at every stage of the student (marked with *).

\subsubsection{Knowledge distillation approaches}
%
%
%
%
Our stagewise approach outperforms the same network
configurations trained without distillation or with the standard distillation approach
(only taking into account the last stage of the cascade).
Learning with hints makes the student MPM-1S
outperform its stagewise distillation counterpart, but 
this is not the case with MPM-2S.
One possible reason
is that given the architecture differences, enforcing feature similarity
prevents the student from  improving the task loss.
In contrast, mimicking the refined predictions at the cascade level using the prediction from
the teacher proves to be effective for distillation.

\subsubsection{Comparison with baseline models}
Knowledge distillation helps boosting the generalization of smaller models.
This is particularly true for MPM-2S which with distillation
reaches the F-Score performance of its deeper (and slower) version MPM-4S baseline.
There is still a small performance gap between the CPM-2S teacher and learner performance
(0.03 in F-Score), but the later runs  10 times faster and is 101 times lighter.
Interestingly, the same conclusions can be drawn from the distillation results obtained on the CMU-Panoptic dataset,
with the MPM-2S improving by 0.1 its F performance using the same distillation approach.

\section{Conclusion}
\label{sec:conclusions}
We have investigated and applied different techniques to achieve fast body
landmark detection in multi-person pose estimation scenarios.
We propose to use depth images in combination with efficient CNN to maintain
the trade-off between speed and accuracy.
Our approach relies on efficient instantiations of the pose machines
with stacked regressors and
employing residual modules, SqueezeNets and MobileNets designs.
In a set of experiments, we have shown that leveraging on knowledge distillation we can
boost the performance of our lightweight models, running at 112.3 FPS with MobileNets with
small performance loss.
We employ and validate the use of synthetic depth images to cope with the lack of
training data and investigated domain adaptation techniques to cope with missing body
landmarks in real data.
Our study suggest that the fusion of sensor data  with synthetic depth images aids the models'
generalization capabilities on real data.

\vspace*{-1mm}
\section*{Acknowledgments}
\vspace*{-1mm}

This work was supported by the European Union under the EU Horizon 2020 Research
and Innovation Action MuMMER (MultiModal Mall Entertainment Robot), project ID 688147,
as well as the Mexican National Council for Science and Technology (CONACYT) under the
PhD scholarships program.

\vspace*{-1mm}

%
%
%
%
%
%
%

%
%
%
%
%
%
%

%
%
%
\vspace{-10mm}

\begin{IEEEbiography}
	[\vspace{-0.75cm}{\includegraphics[width=0.70in,clip,keepaspectratio]{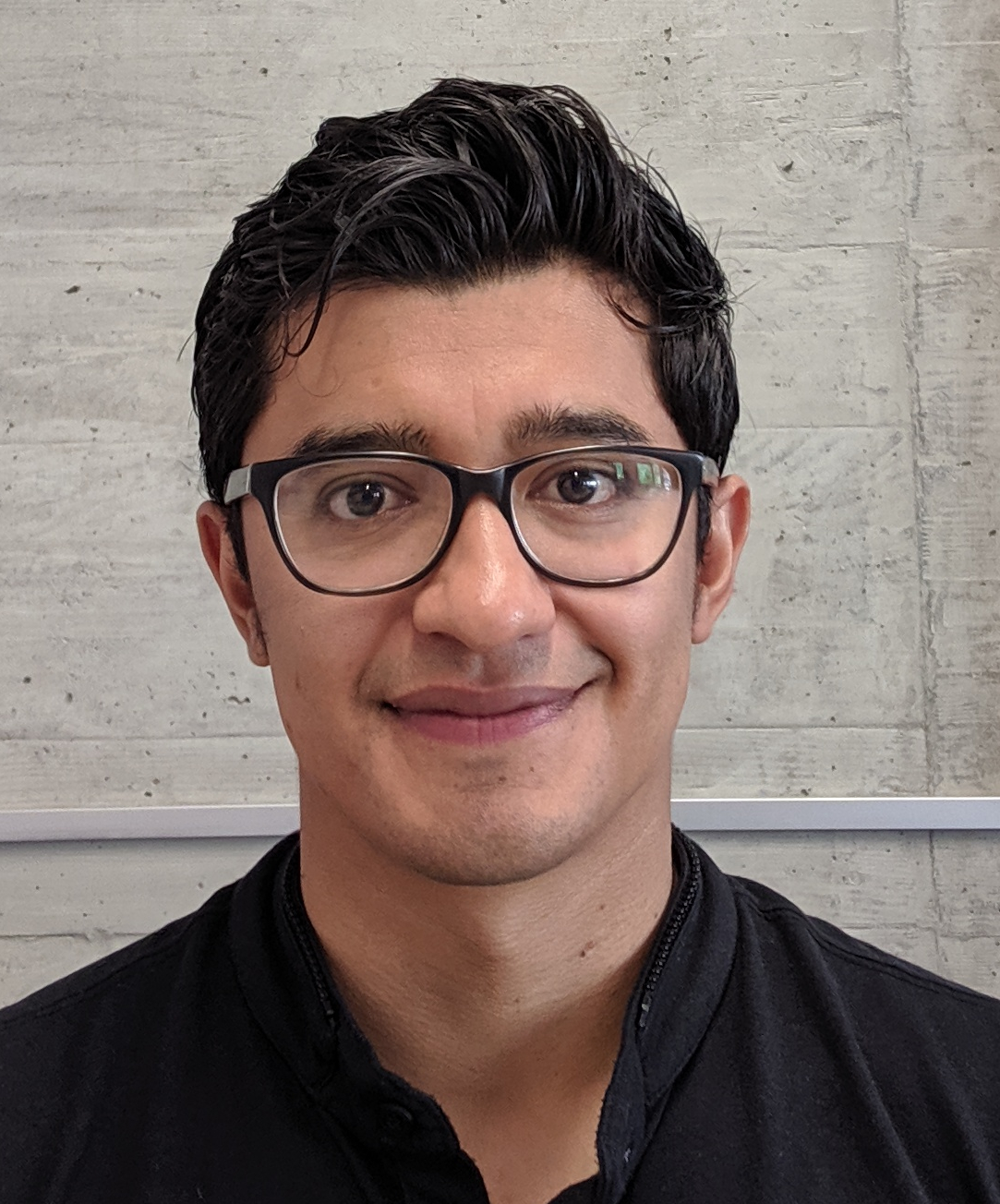}}]{Angel Mart\'inez-Gonz\'alez}

is a research assistant at the Idiap Research Institute and a Ph.D. student at EPFL, Switzerland.
He received his M.Sc. in Computer Science from the Mathematical Research Center (CIMAT) in Guanajuato, Mexico.
%
%
His research interests are machine learning, computer vision and human-computer interaction.

\end{IEEEbiography}

\vspace{-20mm}

\begin{IEEEbiography}
    [\vspace{-0.75cm}{\includegraphics[width=0.70in,clip,keepaspectratio]{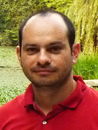}}]{Michael Villamizar}
is a postodoctoral researcher at the Idiap Research Institute (Switzerland).
Before, he was a postdoctoral researcher at the Institut de Rob\`otica i Inform\`atica Industrial, CSIC-UPC, in Barcelona (Spain).
He obtained his PhD in computer vision and robotics from the Universitat Polit\`ecnica de Catalunya in 2012.
%
%
His research interests are focused on object detection and categorization, robust visual tracking, and real-time robotics applications.
\end{IEEEbiography}

\vspace{-20mm}

\begin{IEEEbiography}
[\vspace{-0.75cm}{\includegraphics[width=0.70in,clip,keepaspectratio]{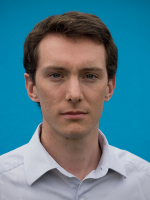}}]{Olivier
  Can\'evet} is a research and development engineer at the Idiap
Research Institute (Switzerland). He received a Ph.D. from the
{\'E}cole Polytechnique F{\'e}d{\'e}rale de Lausanne (Switzerland) in
2016 and an engineering degree from TELECOM Bretagne, (France) in
2012. He takes part in various projects related to computer vision and
machine learning.

\end{IEEEbiography}

\vspace{-20mm}

\begin{IEEEbiography}
    [\vspace{-0.75cm}{\includegraphics[width=0.70in,clip,keepaspectratio]{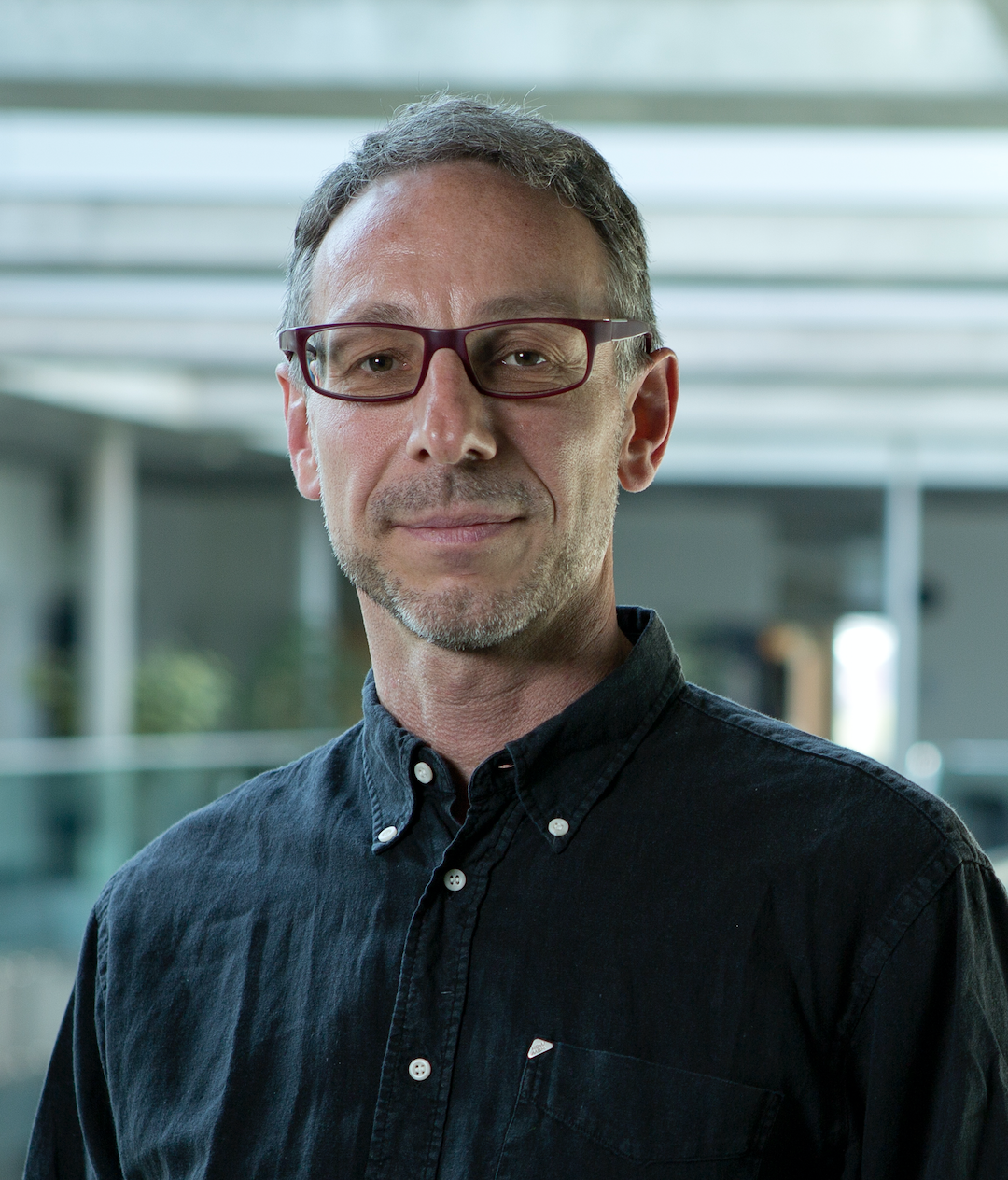}}]{Jean-Marc Odobez}

    (M'03) is the head of the Idiap Perception and Activity Understanding Group and adjunct faculty at the École Polytechnique Fédérale de Lausanne (EPFL).
    He received his PhD from Rennes University/INRIA in 1994.
    He is interested in the design of multimodal perception systems
    for human activity, behavior recognition, human interactions modeling or understanding. 
\end{IEEEbiography}

\end{document}